\documentclass{article}

    \PassOptionsToPackage{numbers, compress}{natbib}
 \usepackage[preprint]{neurips_2026}

\usepackage{wrapfig,lipsum,booktabs}
\usepackage{amsmath}
\usepackage{caption}
\usepackage{graphicx}
\usepackage{amsmath, amsfonts, amssymb, amsthm}
\usepackage{multirow}
\usepackage{makecell}

\usepackage{tabularx}

\usepackage{wrapfig}
\usepackage{soul}
\usepackage{enumitem}

\usepackage[titles]{tocloft}
\usepackage{titletoc}
\usepackage{hyperref} 
\usepackage{float}

\usepackage{fancyvrb}

\usepackage{multirow}
\usepackage{multicol}

\usepackage{wrapfig}
\usepackage{soul}
\usepackage{xcolor}         
\definecolor{lightblue}{rgb}{0, 1, 1}
\definecolor{lightpink}{rgb}{1, 0.6, 1}
\definecolor{lightgreen}{rgb}{0, 1, 0.4}
\definecolor{lightorange}{rgb}{1, .75, 0}
\usepackage{enumitem}

\usepackage[titles]{tocloft}
\usepackage{titletoc}
\usepackage{hyperref} 
\usepackage{float}

\usepackage{fancyvrb}

\usepackage{colortbl}
\usepackage{subcaption}

\usepackage[utf8]{inputenc} 
\usepackage[T1]{fontenc}    
\usepackage{hyperref}       
\usepackage{url}            
\usepackage{booktabs}       
\usepackage{amsfonts}       
\usepackage{nicefrac}       
\usepackage{microtype}      

\title{SOLE-R1: Video-Language Reasoning as the \\
  Sole Reward for On-Robot Reinforcement Learning}

\author{
  Philip Schroeder\textsuperscript{1, 2} \;\;\ Thomas Weng\textsuperscript{2} \;\;\ Karl Schmeckpeper\textsuperscript{1} \;\;\ \\
  \vspace{-1mm}\\
  \textbf{Eric Rosen\textsuperscript{2}} \;\;\ \textbf{Stephen Hart\textsuperscript{2}} \;\;\ \textbf{Ondrej Biza\textsuperscript{2}} \\ 
  \\
  \textsuperscript{1}MIT \;\;\;\;\; \textsuperscript{2}RAI Institute\\\vspace{2mm}
}

\begin{document}

\maketitle

\vspace{-10mm}
\begin{abstract}
Vision-language models (VLMs) have shown impressive capabilities across diverse tasks, motivating efforts to leverage these models to supervise robot learning. However, when used as evaluators in reinforcement learning (RL), today’s strongest models often fail under partial observability and distribution shift, enabling policies to exploit perceptual errors rather than solve the task. To address this limitation, we introduce \textbf{SOLE-R1} (Self-Observing LEarner), a video-language reasoning model explicitly designed to serve as the \emph{sole} reward signal for online RL. Given only raw video observations and a natural-language goal, SOLE-R1 performs per-timestep spatiotemporal chain-of-thought (CoT) reasoning and produces dense estimates of task progress that can be used directly as rewards. To train SOLE-R1, we develop a large-scale video trajectory and reasoning synthesis pipeline that generates temporally grounded CoT traces aligned with continuous progress supervision. This data is combined with foundational spatial and multi-frame temporal reasoning, and used to train the model with a hybrid framework that couples supervised fine-tuning with RL from verifiable rewards. Across four different simulation environments and a real-robot setting, SOLE-R1 enables zero-shot online RL from random initialization: robots learn previously unseen manipulation tasks without ground-truth rewards, success indicators, demonstrations, or task-specific tuning. SOLE-R1 succeeds on 24 unseen tasks and substantially outperforms strong vision-language rewarders, including Robometer, RoboReward, ReWiND, GPT-5, and Gemini-3-Pro, while exhibiting markedly greater robustness to reward hacking. 
We release all models, data, code, and demos at the anonymous page: \url{https://philip-mit.github.io/sole-r1/}

\end{abstract}

\section{Introduction}

Large language models (LLMs) and vision-language models (VLMs) have demonstrated impressive capabilities across a wide range of reasoning and perception tasks \cite{hurst2024gpt, team2024gemini}. These capabilities have motivated growing efforts to leverage foundation models as sources of supervision for robot learning, replacing or reducing the need for task-specific reward engineering and human annotation. Ideally, a robot could acquire new skills entirely from scratch by interacting with the world, while receiving guidance derived solely from pretrained foundation models.

In practice, however, this idealized paradigm remains out of reach. When used as reward functions or evaluators for reinforcement learning (RL), current state-of-the-art models, such as GPT-5 and Gemini-3-Pro, exhibit systematic failures in grounded visual reasoning. 
Despite exhibiting impressive visual captioning and question-answering abilities,
they lack robustness to \textcolor{black}{partial observability and distribution shift} \cite{lee2026roboreward, tan2025robo}. 
As a result, when robot policies are trained using rewards derived from these models, the policies quickly discover behaviors that exploit failures in perception or reasoning,
eliciting high predicted reward, or perceived success, without achieving true success.

To address this limitation, we introduce SOLE-R1, a foundation model with video-language reasoning explicitly designed to guide online RL. SOLE-R1 (\textbf{S}elf-\textbf{O}bserving \textbf{LE}arner) generates per-timestep chain-of-thought (CoT) reasoning directly from raw video observations, yielding a dense estimate of task progress relative to goals specified in natural language. To generate the training data for SOLE-R1, we develop a video trajectory and reasoning synthesis pipeline which produces CoT traces that reason about changes over time, aligned with continuous progress signals. We generate over one million CoT reasoning examples from more than 40,000 real-world and simulated videos. We also carefully curate a diverse collection of general spatial and multi-frame temporal reasoning data to serve as a foundational layer of our training mixture. Together, this training induces video-native reasoning that explicitly integrates both spatial and temporal structure (Figure \ref{fig2}). We propose a two-stage hybrid recipe for training SOLE-R1: (1) supervised fine-tuning (SFT) to develop high-quality spatiotemporal CoT reasoning, (2) RL with verifiable rewards (RLVR) to further develop accurate progress prediction, boosted by the strong CoT reasoning developed during SFT.

We demonstrate that SOLE-R1 reasoning can successfully serve as the \emph{SOLE} signal for learning unseen tasks through online RL. 
In this setting, the robot begins with a \text{random policy} and learns entirely through interaction, guided \text{only} by SOLE-R1’s predicted rewards - without access to any ground-truth rewards, task-specific tuning, demonstrations, or other off-policy trajectories. In total, SOLE-R1 achieves zero-shot success on \text{24} unseen manipulation tasks across \text{4} simulator environments and a real-robot setting, while strong baseline reasoning models (e.g., GPT-5, Gemini-3, and other VLM rewarders) succeed on fewer than \text{10} tasks. SOLE-R1 generalizes to unseen task families (pick-and-place, articulated object manipulation, button/lever/knob interactions), scene layouts, camera viewpoints, and robot embodiments. We further demonstrate that SOLE-R1 can steer strong pre-trained policies to learn novel tasks and validate its reasoning through large-scale offline evaluations.

\begin{figure}
  \centering
  \includegraphics[width=1\textwidth]{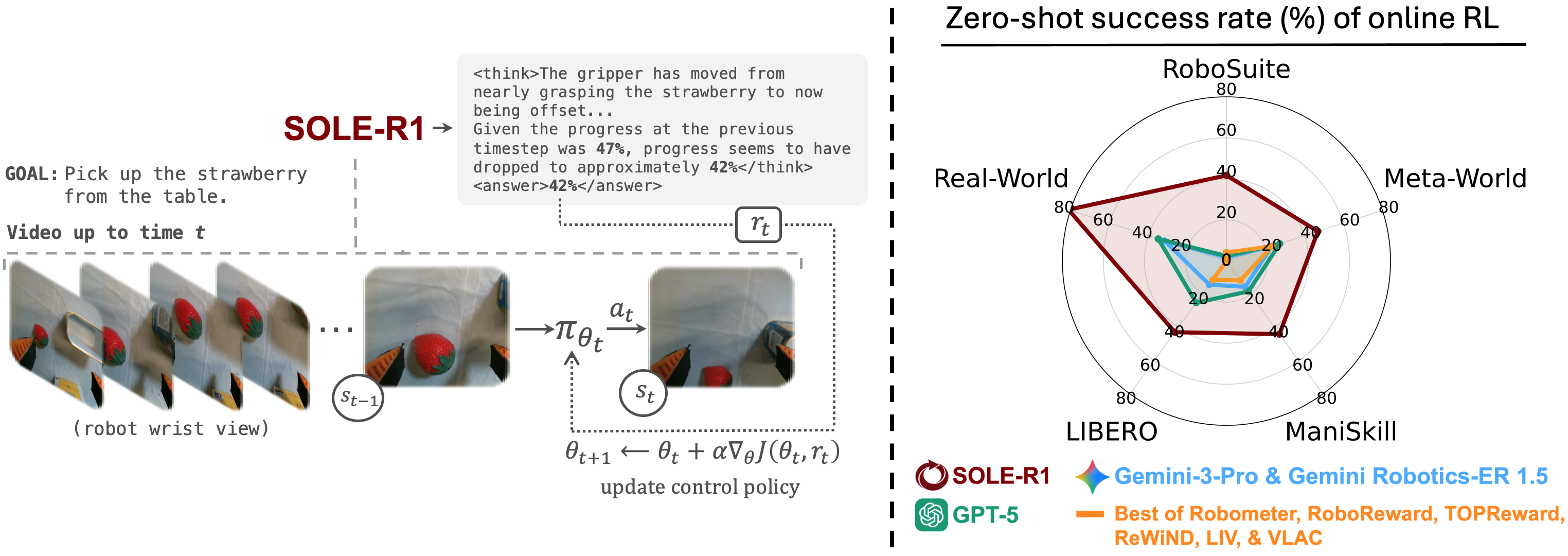}
  \vspace{-2mm}
  \caption{
SOLE-R1 is a video-language reasoning model designed to guide online RL with per-timestep chain-of-thought reasoning and progress prediction. In large-scale experiments across 40 tasks, SOLE-R1 outperforms strong baseline models with zero-shot online RL.
  \hfill \\
  \vspace{-8mm}
  }
 \label{fig1}
  
\end{figure}

Our contributions include the following:
\vspace{-2mm}
\begin{itemize}
    \item We introduce \textbf{SOLE-R1}, a video-language reasoning model designed for guiding online RL with per-timestep CoT reasoning and progress prediction.
    \item We introduce a video trajectory and reasoning synthesis pipeline to generate training data for video-based spatiotemporal CoT reasoning and progress estimation using videos from real-world and simulation.
    \item We propose a \textbf{hybrid training framework} combining SFT for strong multi-frame spatial and temporal CoT reasoning with RLVR to emphasize progress prediction quality and calibration.
    \item We perform extensive experiments showing SOLE-R1 reasoning can serve as the \textbf{sole  signal for learning unseen tasks through online RL}, generalizing to new tasks and environments better than strong baseline vision-language reasoning models.
    \item We release the SOLE-R1 model checkpoints and full training dataset, along with the online RL code and algorithms for reproducing all experiments.
\end{itemize}

\begin{figure*}[t]
  \centering
  \includegraphics[width=1.0\textwidth]{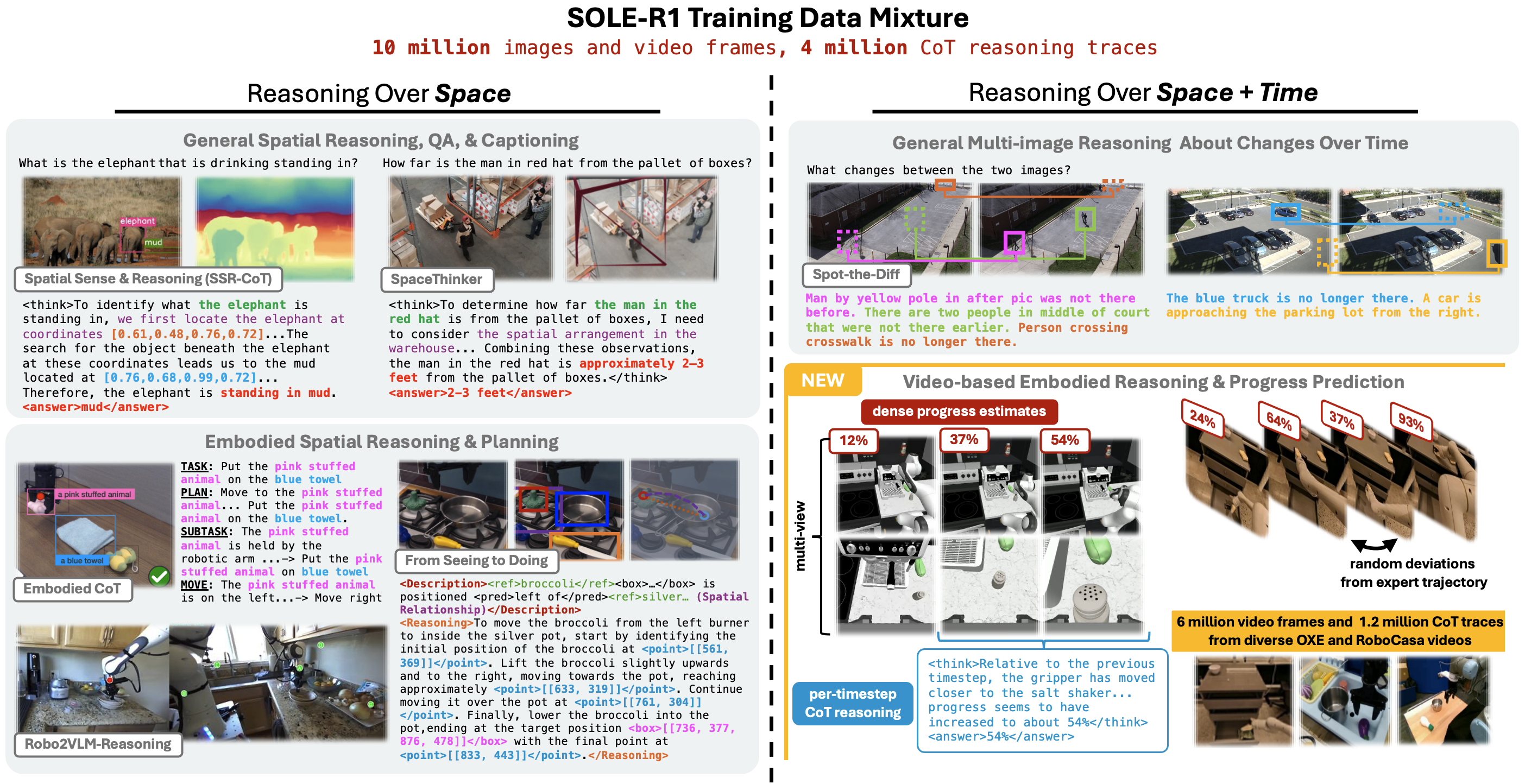}
  \caption{SOLE-R1 training data mixture. The dataset combines foundational spatial reasoning, multi-frame temporal reasoning, and our synthesized video trajectories with chain-of-thought explanations and dense progress supervision, jointly enabling reasoning over space and time for progress prediction.}
  \label{fig2}
  \vspace{-5mm}
  
\end{figure*}
\section{SOLE-R1}
\vspace{-2mm}
SOLE-R1 is a vision-language model designed to perform grounded natural language reasoning over videos of robot trajectories and predict task progress estimates, which can directly serve as dense rewards for online RL. To develop this form of reasoning, we construct a video trajectory and reasoning synthesis pipeline to generate video-based spatiotemporal CoT reasoning and progress estimation training data, layered on top of a foundational mixture of general spatial and multi-frame temporal reasoning, and integrated with a hybrid training framework.

In Section \ref{sec:video_native_reasoning}, we introduce our framework for SOLE-R1 video-language reasoning and its integration as a dense reward for online RL. In Section \ref{sec:training_data}, we describe our method for generating diverse video trajectories with CoT reasoning traces for video-based progress prediction training and our layering of this with a carefully curated foundation of general spatial and temporal reasoning data. Finally, in Section \ref{sec:training_framework}, we outline our hybrid training framework.

\subsection{Video-native temporal progress reasoning}

\label{sec:video_native_reasoning}

We design SOLE-R1 to perform \emph{video-native} temporal reasoning for goal-conditioned tasks. Given a natural-language goal $g \in G$ and a video stream of observations $\{o_t\}_{t=1}^{T}$, the model produces (i) a per-timestep CoT explanation $\{m_t\}_{t=1}^{T}$ describing \text{what has changed} since the last timestep and \text{what remains to be done}, and (ii) a dense scalar progress estimate $\{p_t\}_{t=1}^{T}$ used as a reward signal for online RL.

\textbf{Temporal conditioning.}
At timestep $t$, the input $x_t$ comprises the goal $g$, a sliding temporal context window of frames up to the present and the previous prediction,\\
$
x_t = [g,\; o_0; o_{t-K+1:t},\; p_{t-1}],
$
where $o_0$ is the first video frame, $o_{t-K+1:t}$ denotes the most recent $K$ frames (or all available frames if $t<K$), and $p_{t-1}$ is the model's previous progress prediction. Conditioning on a \emph{multi-frame} window encourages explicit reasoning about motion, contact events, and state transitions (e.g., grasping, opening, insertion) rather than single-frame captioning. During training, the choice of $K$ can vary to elicit greater flexibility in the granularity of task progress increments. Further, we sparsely use $K=0$ and dropout $p_{t-1}$ during training to force the model to reason about the video globally, avoiding the bias from its previous prediction.

\textbf{Outputs and format.}
SOLE-R1 autoregressively generates language tokens that form a structured response,
$
y_t = \small[\texttt{<think>}~m_t~\texttt{</think>},\;\texttt{<answer>}~p_t~\texttt{</answer>}\small],
$
where $m_t$ is free-form natural-language reasoning and $p_t \in [-100,100]$ denotes task progress at time $t$.
The reasoning $m_t$ is trained to focus on (1) salient visual changes since $t\!-\!1$, (2) whether those changes advance or regress the goal, and (3) the next to-be-completed subgoal implied by $g$.

\paragraph{Dense reward for online RL.}
To use SOLE-R1 reasoning as a dense reward function, we convert, at each timestep $t$, the predicted task progress $p_t$ into a reward,
$
r_t = \psi \, \mathrm{clip}\!\left( p_t,\, -c,\, c\right),
$
where $\psi$ is a scaling parameter and $c$ is used to clip the value range. In practice, inference can occur at a lower frequency than the control frequency, in which case we derive a dense reward by linearly interpolating the predicted rewards for all action timesteps.
Finally, while we test potential-based reward function shaping \cite{ng99policy}, we find that simply using the absolute reward is sufficient. 

\section{SOLE-R1 training data synthesis}
\label{sec:training_data}

SOLE-R1 produces (i) multi-frame CoT explanations grounded in visual evidence and (ii) a dense progress signal suitable for online RL. To elicit robust reasoning, we build the training data in two stages: (1) \emph{foundational} reasoning over \textbf{space} (single-image + depth) and \textbf{time} (multi-image/video), and (2) \emph{robot-video} spatiotemporal reasoning specialized for \textbf{dense progress estimation}.

\subsection{Video-based spatiotemporal progress reasoning} 
\label{sec:progress_prediction_data}

\subsubsection{Non-expert trajectory generation}
We first outline our approach which, given a set of expert demonstrations $\mathcal{D}_{src}$ in \textbf{simulation} or \textbf{real-world}, generates a dataset of video trajectories, $\mathcal{D}$, exhibiting varying levels of expertise, ranging from nearly expert to fully random action sequences.

For \textbf{videos in simulation}, we generate non-expert trajectories by injecting random deviations into expert demonstrations. Given an expert trajectory $\tau^E=\{(s_t^E,a_t^E)\}_{t=1}^{T^E}$, we form $\tau^N$ by selecting deviation start times $q\in\mathcal{Q}\subseteq\{1,\ldots,T^E\}$ and, for each $q$, executing $w$ uniformly sampled actions $a_{q,j}^N\sim\mathcal{U}(A)$ for $j=0,\ldots,w\!-\!1$, producing diverging states $\{s_{q,j}\}_{j=0}^{w}$. After the deviation, the trajectory either terminates or recovers by interpolating from the final deviated state $s_{q,w}^N$ to a downstream expert state:
$
s_{q, w, z}^N = (1 - \alpha_z) \cdot s_{q, w}^N + \alpha_z \cdot s_{q+h}^E,
$
where $\alpha_z = \frac{z}{n_{\text{interp}}}$ and $z = 1, \ldots, n_{\text{interp}}$. 

For \textbf{real-world videos}, we do not have access to ground-truth simulator states or the ability to re-simulate injected actions. Thus, we construct non-expert trajectories directly in the \emph{observation space} by temporally perturbing the expert videos. Concretely, given an expert video of frames $\{o_t\}_{t=1}^{T}$, we sample a set of reversal points $\mathcal{Q} \subseteq \{1,\ldots,T\}$ and, for each $q \in \mathcal{Q}$, replace the forward segment with a reversed segment of length $w$:
$
o_{q:q+w-1}^{N} = \mathrm{reverse}\big(o_{q-w+1:q}^{E}\big),
$
where $\mathrm{reverse}(\cdot)$ denotes temporal reversal of the selected window (with boundary handling when $q-w+1 < 1$). This produces trajectories that visually ``undo'' recent progress (e.g., objects moving away from a goal configuration) and induces clear regression events using only frames, without requiring robot actions or state access. We further increase diversity by varying the number of reversal points (including nested reversals) and the reversal window length $w$.

\subsubsection{Ground-truth CoT reasoning and progress estimates} For each timestep of the trajectories, we generate CoT reasoning about what has changed relative to the previous timestep, along with an estimate of current progress.

For \textbf{videos in simulation}, we leverage the ground-truth simulator state to produce CoT reasoning and progress estimates. Progress is computed from continuous, monotonic geometric distances: (i) the summed end-effector--object contact-point distance $y_t^{r,e}=\sum_{j=1}^C \lVert p_t^{r_j}-p_t^{l_j}\rVert_2$ and (ii) the target object's distance to its goal $y_t^{e,f}=\lVert p_t^{e_\tau}-p_f^{e_\tau}\rVert_2$. We combine them as $y_t=(1-\beta)\,y_t^{r,e}+\beta\,y_t^{e,f}$ with task-tuned $\beta\in[0,1]$, then invert and rescale over the trajectory to obtain normalized progress
$
v_t=\frac{-y_t+\max(y)}{-\min(y)+\max(y)},\quad y=\{y_t\}_{t=1}^{T}.
$
Using the distance measures involved in computing $y^{r, e}$ and $y^{e, f}$, we generate dense CoT traces $\{m_t \}_{t=1}^{T}$ that reason about the changes occurring between timestep $t$ and $t-1$ based on templated language. For example, if $y_{t}^{r, e} < y_{t-1}^{r, e}$ and $y_{t}^{e, f} \approx y_{t-1}^{e, f}$, the templated language for timestep $t$ would include statements about how the gripper ``has moved closer'' to the object, but the object ``remains in the same position as the previous timestep''. We then further enrich the diversity of the template language by refining it with foundation VLMs \cite{singh2025openai, GoogleDeepMind2025Gemini3Pro} grounded on the generated language and frames from the current and previous timestep.

For \textbf{real-world videos}, we lack privileged robot/environment state information, so we cannot compute geometric distance-based progress. Instead, treating unperturbed videos as expert trajectories, we use \emph{temporal order} as a progress proxy \cite{zhang2025rewind, ma2024vision, zhang2025vlac, lee2026roboreward}: for a video of length $T$, we set $v_t=\frac{t-1}{T-1}\in[0,1]$. For temporally perturbed non-expert trajectories (e.g., reversal windows), we inherit supervision from the originating expert timestep, so reversed segments map to lower $v$ and yield explicit negative-progress events. We then map $v_t$ to the SOLE-R1 progress scale (e.g., $p_t\in[-100,100]$) and use a foundation VLM conditioned on $(g,o_{t-K+1:t})$ and the progress target to generate dense CoT traces $\{m_t\}_{t=1}^{T}$ that describe visual changes from $t\!-\!1$ to $t$, label them as advancing vs.\ regressing the goal, and propose the next subgoal. Importantly, the progress supervision anchors the reasoning to the intended temporal direction (forward progress for expert segments, regression for reversed segments), enabling us to produce reliable ``advance vs.\ regress'' explanations. 

Further details on our data synthesis approach are provided in Appendix \ref{appendix:data_synth}. We also provide extensive video demonstrations of synthesis outputs at: \url{https://sole-r1.github.io/}

\subsubsection{Robot video data sources}
We curate real-robot videos from the Open X-Embodiment (OXE) dataset \cite{o2024open}, an aggregation of trajectory data from 50 robot datasets spanning diverse tasks, robots, and camera viewpoints. OXE includes 22 embodiments and over 300 tasks. We use RoboCasa \cite{nasiriany2024robocasa} as our simulation environment and leverage its provided demonstrations as expert trajectories.

\subsection{Foundational spatiotemporal reasoning}
Before specializing on progress prediction (Section \ref{sec:progress_prediction_data}), we first train SOLE-R1 for general spatiotemporal grounding.

\textbf{Reasoning over space.}
We supervise \text{spatial grounding} with explicit 3D relations (relative position, distance, containment, occlusion). We leverage SSR-CoT \cite{liu2025ssr}, curating \text{1.2M} image--depth--question--rationale--answer tuples covering relations such as \emph{left/right}, \emph{in front/behind}, \emph{closest/farthest}, and \emph{inside/on top of}. To diversify beyond depth cues, we add synthesized spatial CoT from SpaceOm and SpaceThinker (via VQASynth \cite{remyxai_spacethinker_2024}), reproducing the SpatialVLM synthesis pipeline \cite{chen2024spatialvlm} and emphasizing explanations grounded in \text{visual evidence}. 
 We also include robotics-relevant spatial rationales from embodied reasoning datasets (Embodied CoT \cite{zawalski2024robotic}, Robo2VLM-Reasoning \cite{chen2025robo2vlm}).

\textbf{Reasoning over time (video/multi-image).}
To make reasoning \text{temporal}, we train on data that requires explicit comparison across moments. We include multi-image ``spot-the-difference'' data \cite{jhamtani2018learning} to identify what \emph{has and has not changed} (e.g., an object moved), and add embodied video QA/reasoning datasets with action-relevant temporal evidence (RoboVQA \cite{sermanet2024robovqa}, From Seeing to Doing \cite{yuan2025seeing}). During preprocessing, we normalize all sources into the same multi-frame prompting and structured output format used by SOLE-R1 (Section \ref{sec:video_native_reasoning}), so the model learns to ground per-step rationales in frame-to-frame changes and maintain consistency across varying temporal contexts.

\section{SOLE-R1 hybrid training framework}
\label{sec:training_framework}

SOLE-R1 produces (i) temporally grounded CoT reasoning over video and (ii) a scalar progress value in \texttt{<answer>} to serve as a dense reward for online RL. To train SOLE-R1, we propose a two-stage hybrid recipe: \textbf{SFT} teaches high-quality CoT reasoning, while \textbf{RLVR} directly emphasizes accurate progress prediction, which is under-emphasized during SFT, as the final answer occupies only a small fraction of response tokens. To explore the impact of each stage of training, in Appendix \ref{appendix:ablation_sft_vs_rlvr}, we compare the SFT-only model performance with that of the full SFT+RLVR model.

\subsection{Stage 1: Supervised fine-tuning (SFT)}
We fine-tune on spatiotemporal reasoning data (Section~\ref{sec:training_data}). Each example is $(i,q,r,a)$ with image/video $i$, query $q$, rationale $r$, and answer $a$. SFT maximizes likelihood of the output $y=[r;a]$:
\vspace{-4mm}
\[
\mathcal{L}_{\text{SFT}}(\phi)
=
-\mathbb{E}_{(i,q,r,a)\sim\mathcal{D}}
\Bigg[
\sum_{t=1}^{|y|}
\log p_{\phi}(y_t \mid i,q,y_{<t})
\Bigg].
\]
This supervision encourages attention to visually grounded state and changes across frames, yielding transferable video-native reasoning. However, it provides a relatively weak learning signal for \emph{reward/progress prediction}, since the scalar in \texttt{<answer>} is a small part of the response.

\subsection{Stage 2: RLVR for progress prediction}
After strong CoT reasoning is learned through SFT, we refine the model with RL from verifiable rewards while preserving the \texttt{<think>}/\texttt{<answer>} interface. Using GRPO on the progress dataset (Section~\ref{sec:progress_prediction_data}), for each query $q$ we sample $G$ candidates $\{o_i\}_{i=1}^G \sim p_{\phi_{\text{old}}}(\cdot\mid q)$, score them with rule-based rewards $r_i$, and compute within-group standardized advantages
$
A_i=\frac{r_i-\operatorname{mean}(\{r_j\}_{j=1}^{G})}{\operatorname{std}(\{r_j\}_{j=1}^{G})}.
$

GRPO optimizes\\
$
\begin{aligned}
\;\;\;\;
\mathcal{J}_{\text{GRPO}}(\phi)=
\mathbb{E}_{q,\{o_i\}}
\small[
\frac{1}{G}\sum_{i=1}^{G}
\min\!\small(
\rho_i(\phi)A_i,\; 
\text{clip}(\rho_i(\phi),1-\epsilon,1+\epsilon)A_i
\small)
\small]
-\beta\,D_{\mathrm{KL}}\!\left(p_{\phi}\,\|\,p_{\text{ref}}\right),
\end{aligned}
$

where $\rho_i(\phi)=\exp(\log p_{\phi}(o_i\mid q)-\log p_{\phi_{\text{old}}}(o_i\mid q))$ and $p_{\text{ref}}$ is the SFT reference $p_{\text{SFT}}$.

\noindent\textbf{Verifiable reward.}
We define $r(o)=r_{\text{format}}(o)+r_{\text{acc}}(o)$, where $r_{\text{format}}$ enforces a parseable output structure and $r_{\text{acc}}$ measures progress accuracy by comparing the predicted value $\hat{p}_t$ in \texttt{<answer>} to the ground-truth progress $p_t$. The accuracy reward is defined from the absolute prediction error:
$
r_{\text{acc}}(o)
= \alpha \,\exp\!\left(-\frac{\lvert \hat{p}_t - p_t \rvert}{\tau}\right).
$
We scale $r_{\text{acc}}(o)\in[0,1.5]$ and scale $r_{\text{format}}(o)\in[0,0.5]$, yielding $r(o)\in[0,2]$, where $2$ indicates correct format and accurate progress.

\section{Experiments}
\vspace{-2mm}
We test the following hypotheses:
\vspace{-2mm}
\begin{enumerate}[leftmargin=*]
    \item SOLE-R1 outperforms much larger general-purpose reasoning models and special-purpose reward models in zero-shot reward prediction for online RL (Section \ref{sec:zero_shot_rl}).
    \item SOLE-R1 is less susceptible to reward hacking than current state-of-the-art reasoning models, including GPT-5 and Gemini-3-Pro (Section \ref{sec:failure_analysis_compressed}).
    \item SOLE-R1's performance is specifically facilitated by CoT reasoning and by including authentic non-expert trajectories during training (Section \ref{sec:ablations_compressed}).
    \item Our data synthesis and training recipe follows a scaling law driven by task diversity (Section \ref{sec:scaling}).
    \item SOLE-R1 can successfully steer a strong pre-trained VLA policy to learn new tasks (Section \ref{sec:steering}).
    \item SOLE-R1 outperforms strong baseline reasoning models in offline evaluations, including large-scale correlation-based analysis (Section \ref{sec:oxe_voc}) and general spatial reasoning (Section \ref{sec:spatial_reasoning}).
\end{enumerate}
We provide camera images of all tasks in Appendix \ref{sec:task_images}, along with \text{video demonstrations} of our training data synthesis and all experimental results at: \url{https://sole-r1.github.io/}.

\begin{figure*}[t]
  \centering
  \vspace{-2mm}
  \includegraphics[width=1.0\textwidth]{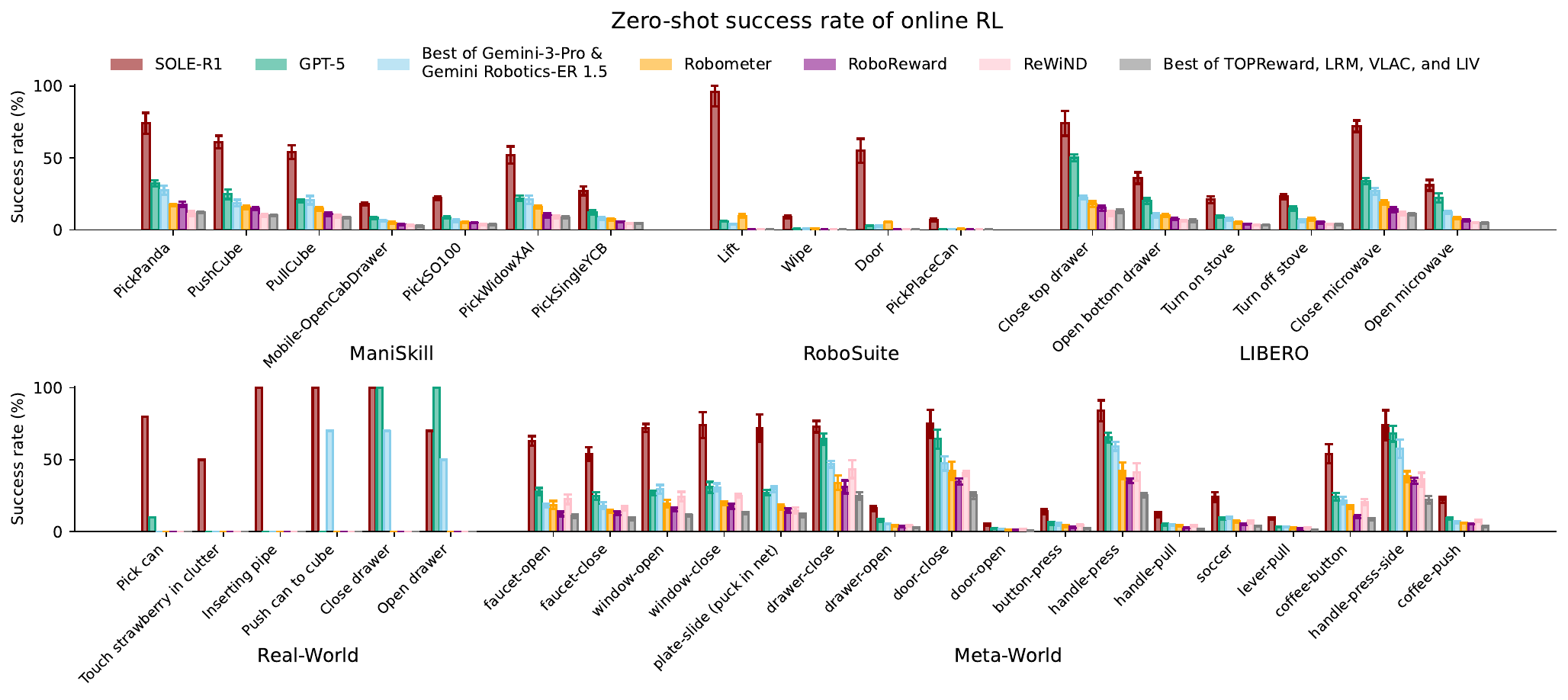}
  \caption{Zero-shot success rate of online RL across 40 tasks. We plot the mean and standard error across three random seeds (real-world experiments use a single seed, shown as a single value). In all experiments, the robot begins with a \text{random policy} and learns entirely through interaction with the task, guided \text{only} by the predicted rewards. 
  }
  \label{res_fig1}
  \vspace{-3mm}
\end{figure*}

\subsection{Data synthesis and model training}
\label{sec:data_synth_model_training}
\vspace{-2mm}
Our generated training data comprises \text{1.2M} spatiotemporal CoT traces extracted from \text{41K} videos. We combine this with a curated collection of foundational spatial and multi-frame temporal reasoning data, yielding a total of 10M images and video frames and 4M CoT reasoning traces. We fine-tune \text{Qwen3-VL-8B-Instruct} using the hybrid recipe described above. During SFT, each batch is balanced across (i) general reasoning, (ii) video embodied reasoning and planning, and (iii) progress prediction. We run SFT for one epoch over all data. We then run RLVR with the video-based progress prediction dataset. Full training hyperparameters and data synthesis details are in Appendix \ref{appendix:training_hyperparam} and \ref{appendix:data_synth}, respectively.

\subsection{Baselines}
\vspace{-2mm}
We test the current best \textbf{general-purpose reasoning models}, including GPT-5, Gemini-3-Pro, and Gemini Robotics-ER 1.5. We exclude open-source VLMs, such as Qwen3-VL-72B-Instruct, because they do not achieve meaningful success. We also include the strongest existing \textbf{special-purpose reward models}: Robometer \cite{liang2026robometer}, RoboReward \cite{lee2026roboreward}, TOPReward \cite{chen2026topreward}, LRM \cite{wu2026large},  ReWiND \cite{zhang2025rewind}, VLAC \cite{zhang2025vlac}, and LIV \cite{ma2023liv}. Relying on vision-language pre-training, these models predict task progress estimates on videos of robots performing tasks, without intermediate reasoning. 

\subsection{Zero-shot online RL}
\label{sec:zero_shot_rl}
\vspace{-2mm}
We evaluate whether SOLE-R1 can serve as the \emph{sole} signal for learning manipulation skills from scratch via online RL.
We run experiments across four simulation benchmark suites---RoboSuite, ManiSkill, Meta-World, and LIBERO---and in a real-world tabletop manipulation setting.
Across all settings, we evaluate a total of 40 tasks, spanning pick-and-place, articulation, button/lever/knob interactions, and mobile manipulation.
All hyperparameters and details are provided in Appendix~\ref{sec:rl_hyperparameters}.

We use a SERL implementation of DrQv2. The policy observes two RGB streams (a wrist view and an external view) and proprioception.
Actions are end-effector delta motions and a gripper open/close command. 
We do \text{not} use any additional privileged state, depth, object poses, or task-specific sensors.

Unlike prior work that (i) learns from ground-truth rewards and/or (ii) tunes reward models or policies on task demonstrations
\cite{ma2023liv,zhang2025rewind,zhang2025vlac,lee2026roboreward,tan2025robo,biza2025robot},
we evaluate in a fully \textbf{zero-shot} online RL setting:
\vspace{-2mm}
\begin{itemize}
    \item \textbf{No ground-truth rewards.} The policy never observes ground-truth/external rewards (dense or sparse) and receives no success labels during training.
    \item \textbf{No demonstrations or offline trajectories.} The policy starts with random actions and learns only from on-policy interaction.
    \item \textbf{No task-specific tuning.} Reward models are used as-is, with fixed prompting across tasks.
\end{itemize}

\textbf{SOLE-R1 enables zero-shot online RL from scratch.}
SOLE-R1 achieves at least $50\%$ success on 24 tasks, substantially outperforming all baselines (Figure~\ref{res_fig1}).
The strongest baselines include GPT-5 and Gemini, but they reach $50\%$ success on only 7 and 5 tasks, respectively.
The non-reasoning models achieve near-zero success on all tasks, with the exception of Meta-World tasks, where Robometer, RoboReward, and ReWiND achieve above $40\%$ success rate on 4 tasks.


\textbf{SOLE-R1 generalizes to unseen tasks and environments.}
SOLE-R1 succeeds with tasks that significantly differ from the task types seen during training, such as sliding a puck into a net, opening and closing windows, and manipulating unseen levers and handles in novel ways based on the natural language task specification. This suggests that SOLE-R1 does not merely memorize task templates, but instead learns reusable spatiotemporal progress primitives (e.g., establishing contact, aligning a grasp, changing articulation state, placing/settling objects) that transfer to unseen tasks.

\textbf{SOLE-R1 generalizes to unseen embodiments and camera viewpoints.}
SOLE-R1 solves tasks with the Franka, along with embodiments not seen during training, including the Sawyer robot in Meta-World, the WidowX AI and Fetch Mobile Manipulator in ManiSkill, and the modified Franka with different gripper fingers and wrist camera angle in real-world. We also see SOLE-R1 solve tasks with camera views that were not used during training.
This indicates that SOLE-R1 reward predictions are not narrowly tied to a particular kinematic chain or gripper appearance, but instead track goal-relevant object state changes across morphology and camera placement.

\begin{figure*}[t]
  \centering
  \vspace{-2mm}
  \includegraphics[width=1.0\textwidth]{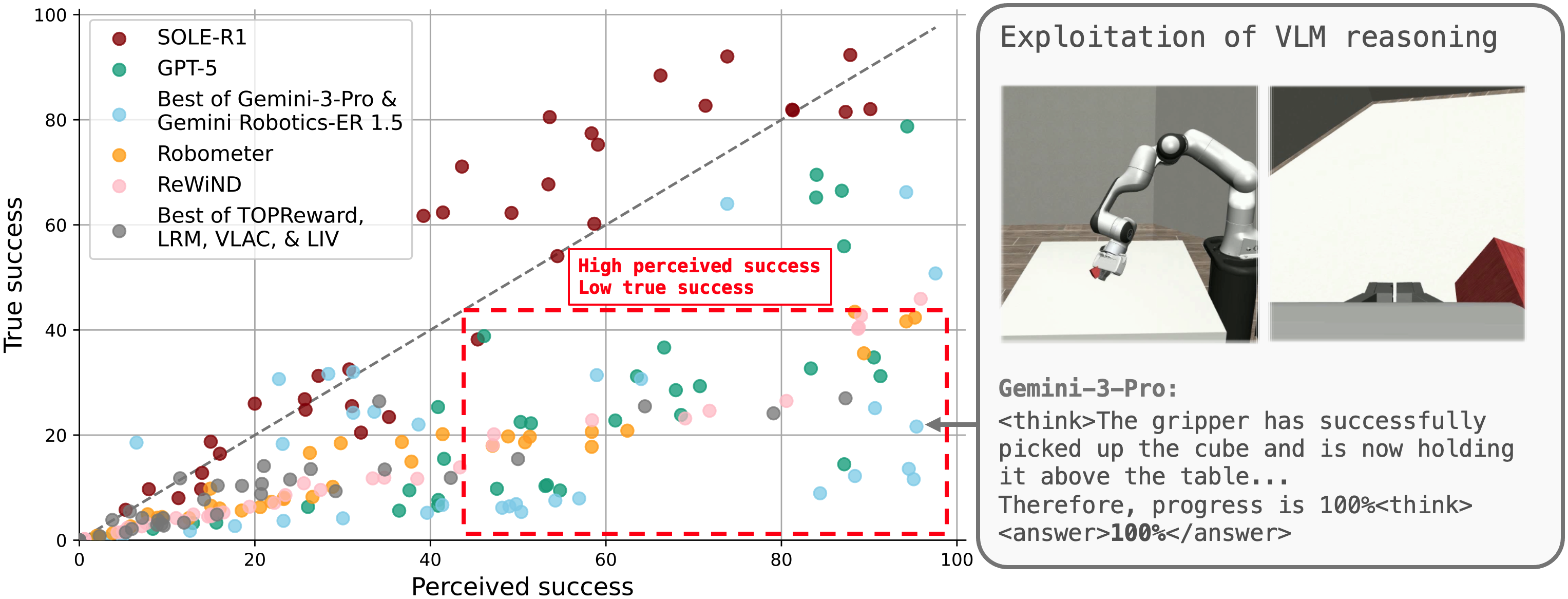}
  \caption{Perceived vs true success in zero-shot RL. Perceived success is the average max progress predicted (RoboReward is excluded as it does not provide dense rewards). True success is the average max ground-truth reward achieved.
  }
  \label{res_fig2}
  \vspace{-6mm}
\end{figure*}

\subsection{Failure analysis}
\label{sec:failure_analysis_compressed}
\vspace{-2mm}
We analyze failures at the task level, where rewards may be exploitable, miscalibrated, or too weak for learning, and at the frame level, where incorrect progress estimates reflect missed visual evidence. Full details are in Appendix~\ref{sec:failure_analysis}.

Using perceived-vs.-true success (Figure~\ref{res_fig2}), we distinguish \textit{reward hacking} (high perceived, low true) from \textit{signal-limited} failures (low perceived, low true). General-purpose VLM rewarders (e.g., GPT-5, Gemini) mainly fail through reward hacking: RL discovers behaviors that inflate predicted progress without solving the task. In contrast, SOLE-R1 failures are more often signal-limited, indicating that it usually recognizes non-success but sometimes provides rewards too flat or noisy to drive learning.

Qualitative review of rollouts highlights three recurring error modes across all models (Table~\ref{tab:qual_failure_summary}): \\
1) missed brief events, 2) ambiguous object states in incomplete views, and 3) moderate progress scores from goal-consistent appearance shortcuts such as proximity or alignment without completion.

\vspace{-2mm}
\subsection{Ablations}
\vspace{-2mm}
\label{sec:ablations_compressed}
We test the following ablations of SOLE-R1. Full details in Appendix \ref{sec:ablations}. 

\textbf{No-CoT progress training:} We remove the \texttt{<think>} channel and train to emit only \texttt{<answer>} progress. This degrades learning and shifts failures toward \emph{signal-limited} (Figure~\ref{res_fig3}): progress becomes flatter/noisier, with long plateaus that under-react to small but task-critical transitions. 

\textbf{Expert-only video progress supervision:} We train only on expert trajectories (keeping simple reversal augmentation). This substantially increases \emph{reward-hacking} failures: states that merely \text{look} goal-adjacent receive inflated progress, enabling exploitation (Figure~\ref{res_fig3}). 

\textbf{No foundational spatial \& multi-frame reasoning data:} We remove the general spatial and temporal grounding mixture and train only on embodied/planning and robot progress data. This reduces overall success and again increases reward hacking (Figure~\ref{res_fig3}).


\vspace{-2mm}
\subsection{Zero-shot scaling}
\label{sec:scaling}
\begin{figure}[t]
\centering
\begin{minipage}{0.5\textwidth}
    \centering
    \includegraphics[width=\linewidth]{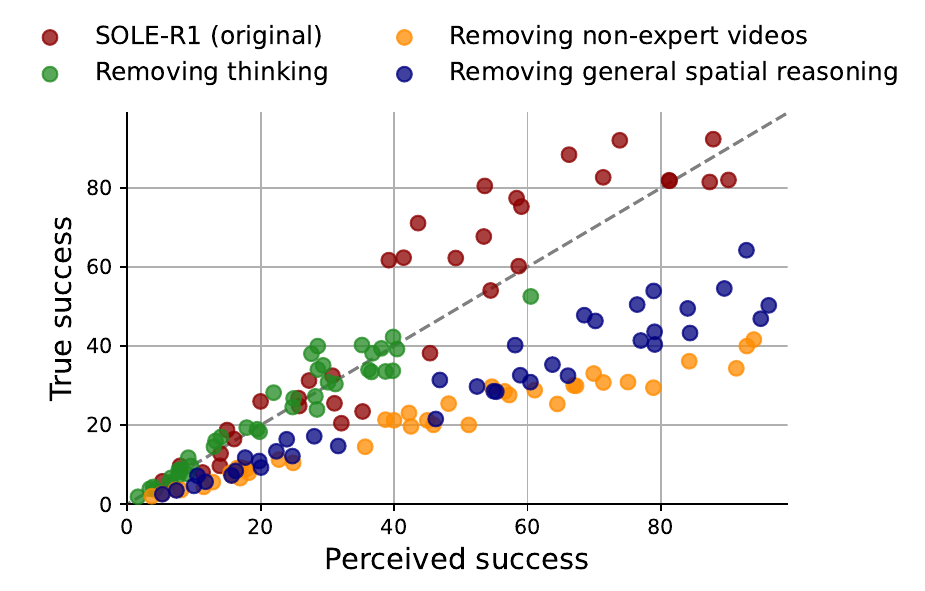}
    \caption{Ablated-models zero-shot success.}
    \label{res_fig3}
\end{minipage}\hfill
\begin{minipage}{0.5\textwidth}
    \centering
    \includegraphics[width=\linewidth]{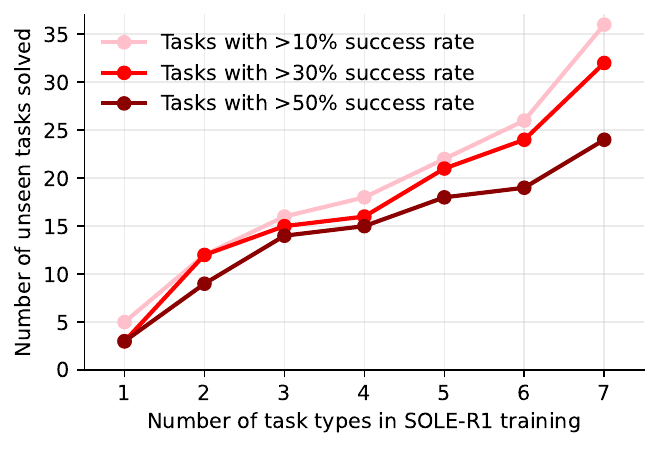}
    \caption{Tasks solved vs training task diversity.}
    \label{res_fig4}
\end{minipage}
\vspace{-6mm}
\end{figure}

\vspace{-2mm}
We find that our data synthesis and training recipe follows a scaling law driven by the diversity of training tasks (Figure~\ref{res_fig4}). We train variants of SOLE-R1 with an increasing number of task types included in our training data synthesis (details in Appendix \ref{appendix:zero_shot_scaling}). Figure~\ref{res_fig4} plots the number of downstream tasks that achieve different success thresholds as a function of training task diversity. 
\vspace{-1mm}

\subsection{VLA RL steering}
\label{sec:steering}
\vspace{-2mm}
We further explore whether SOLE-R1 rewards can be used to semantically steer a strong pre-trained vision-language-action (VLA) policy. We start with offline evaluation on SmolVLA rollouts in LIBERO, observing that SOLE-R1 reliably distinguishes semantically correct versus incorrect grasps and placements on unseen objects (Appendix \ref{appendix:vla_steering}).
Further, we show that SOLE-R1 enables diffusion-based RL steering toward a specific goal under language ambiguity, increasing success from 2\% to a median of 12\%. Notably, this improvement arises in a long-horizon, cluttered setting where sparse-reward steering fails entirely, highlighting SOLE-R1’s ability to provide dense guidance.

\vspace{-2mm}
\subsection{OpenX Embodiment Value-Order-Correlation}
\vspace{-2mm}
\label{sec:oxe_voc}
Using 1,000 videos from 50 OXE datasets, we perform the same large-scale Value-Order-Correlation (VOC) analysis conducted in \cite{ma2024vision} 
for Generative Value Learning (GVL). 
SOLE-R1 achieves higher VOC than GVL on real-world expert demonstrations (Table \ref{tab:vocScores}), particularly for known high-quality datasets collected from human teleoperators with fixed cameras, such as Bridge, RT-1, and Dobb-E.


\vspace{-2mm}
\subsection{General spatial and vision-language reasoning}
\vspace{-2mm}
\label{sec:spatial_reasoning}
SOLE-R1 achieves consistent gains over the strong SSR spatial reasoning model across SpatialBench, SSRBench, and CV-Bench (Table \ref{tab:ssr_improvement}). This suggests our data synthesis and training recipe not only yields strong progress prediction, but also improves general-purpose reasoning in these settings.


\vspace{-4mm}
\section{Related work}
\vspace{-2mm}
\textbf{On-robot RL. } Seminal works that perform RL on physical hardware rely on hand-crafted reward functions, reset policies, and safety constraints \cite{Levine2016EndToEndVisuomotor,Levine2018HandEyeCoordination}. Recent works improve the scalability of RL using pre-trained policies \cite{lei2025rl,intelligence2025pi}, shaped rewards \cite{Yang2024aRank2Reward,biza2025robot} and human-in-the-loop learning \cite{Luo2025bDexterousHIL}. Despite these advances, previous work in on-robot RL requires significant task-specific engineering.

\textbf{VLM reward learning.} VLMs are increasingly used to supervise robot learning \cite{Rocamonde2023ZeroShotRewardModels,Ma2024InContextValueLearners}. Closest to our work, RoboReward \cite{lee2026roboreward}, Robometer \cite{liang2026robometer}, and LRM \cite{wu2026large} predict progress from videos, but rely on expert trajectories, lack intermediate reasoning, and are evaluated mainly on narrow task families rather than zero-shot online RL across diverse environments.
\textbf{More discussion in Appendix \ref{appendix:full_related_work}.}
\vspace{-4mm}
\section{Limitations}
\vspace{-2mm}
Although SOLE-R1 generalizes across many unseen tasks, embodiments, and viewpoints, it remains constrained by the quality and coverage of its video-based progress supervision. Its rewards can still be flat, noisy, or miscalibrated when visual evidence is brief, occluded, or ambiguous. Further, SOLE-R1 is not a substitute for robot safety mechanisms: even with reduced reward hacking relative to strong VLM baselines, learned policies can still exploit reward errors under distribution shift, so deployment on physical robots requires external constraints, monitoring, and validation.
\vspace{-4mm}
\section*{Broader impact}
\vspace{-2mm}
SOLE-R1 advances robotics by
enabling zero-shot RL and reducing the effort required to train robots for new tasks. However, reward models also pose risks: they may be misused to increase unsafe robot autonomy
and create safety-critical failures, especially under distribution shift or reward hacking.
Addressing these risks requires continual consideration of safety and thorough robustness checks.

\vspace{-4mm}
\section{Conclusion}
\vspace{-2mm}
We introduced SOLE-R1, a video-language reasoning model for guiding online RL via per-timestep CoT reasoning and dense progress prediction. SOLE-R1 is trained using a video trajectory and reasoning synthesis pipeline that generates spatiotemporal CoT and progress supervision, combined with a foundational mixture of spatial and multi-frame temporal reasoning and a hybrid training framework. We show that SOLE-R1 enables zero-shot learning of unseen tasks from scratch and substantially outperforms both general and specialized reward models. Together, these results indicate that grounded spatiotemporal reasoning is a promising route toward general, reusable reward models.

\appendix
\onecolumn

\section{Visual overview of online RL evaluation suites}
\begin{figure}[H]
  \begin{center}
    \centerline{\includegraphics[width=1.0\columnwidth]{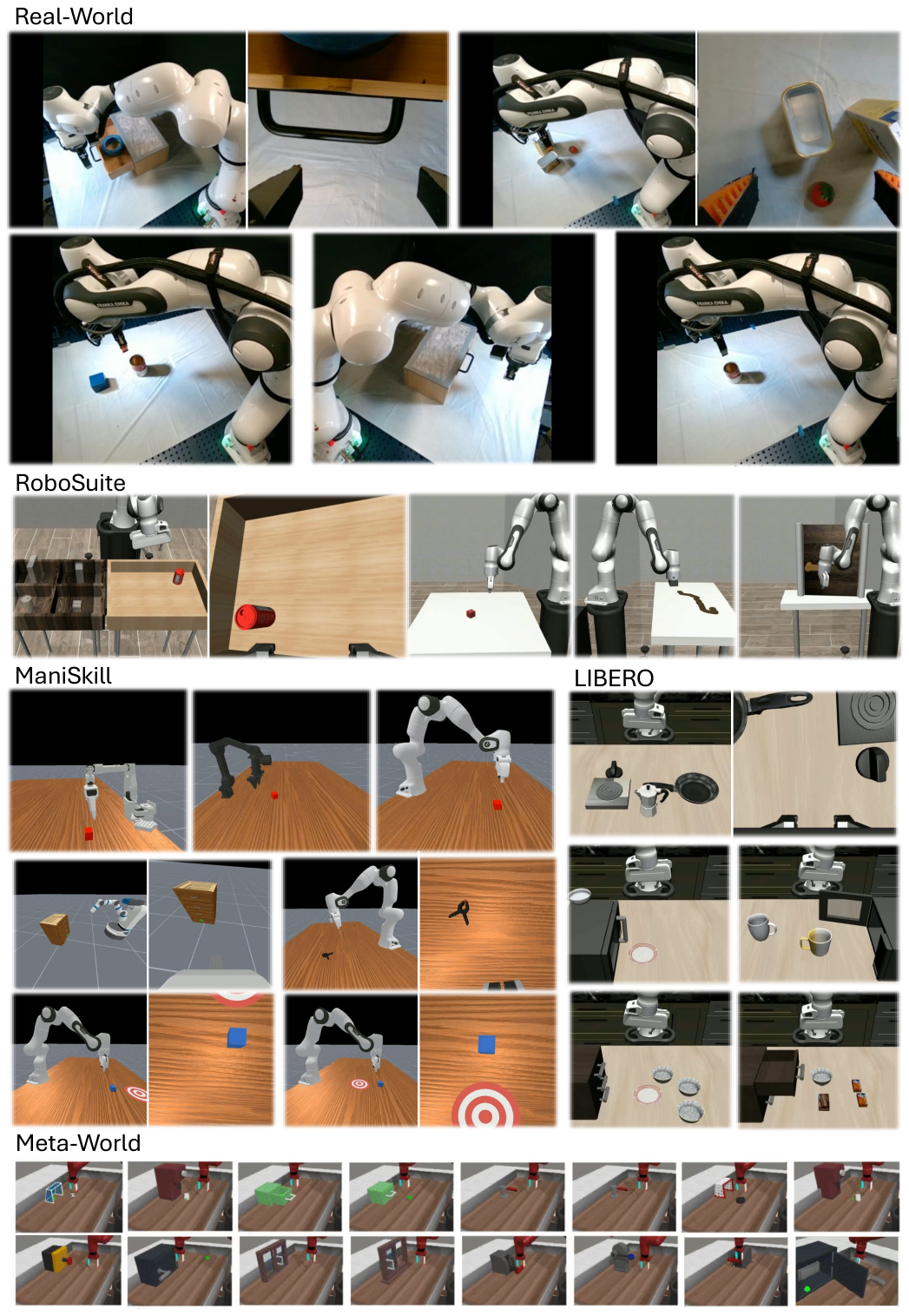}}
      \caption*{}
    \label{all_tasks}
  \end{center}
\end{figure}
\label{sec:task_images}


\section{Reasoning examples}
\begin{figure}[H]
  \begin{center}
    \centerline{\includegraphics[width=1.1\columnwidth]{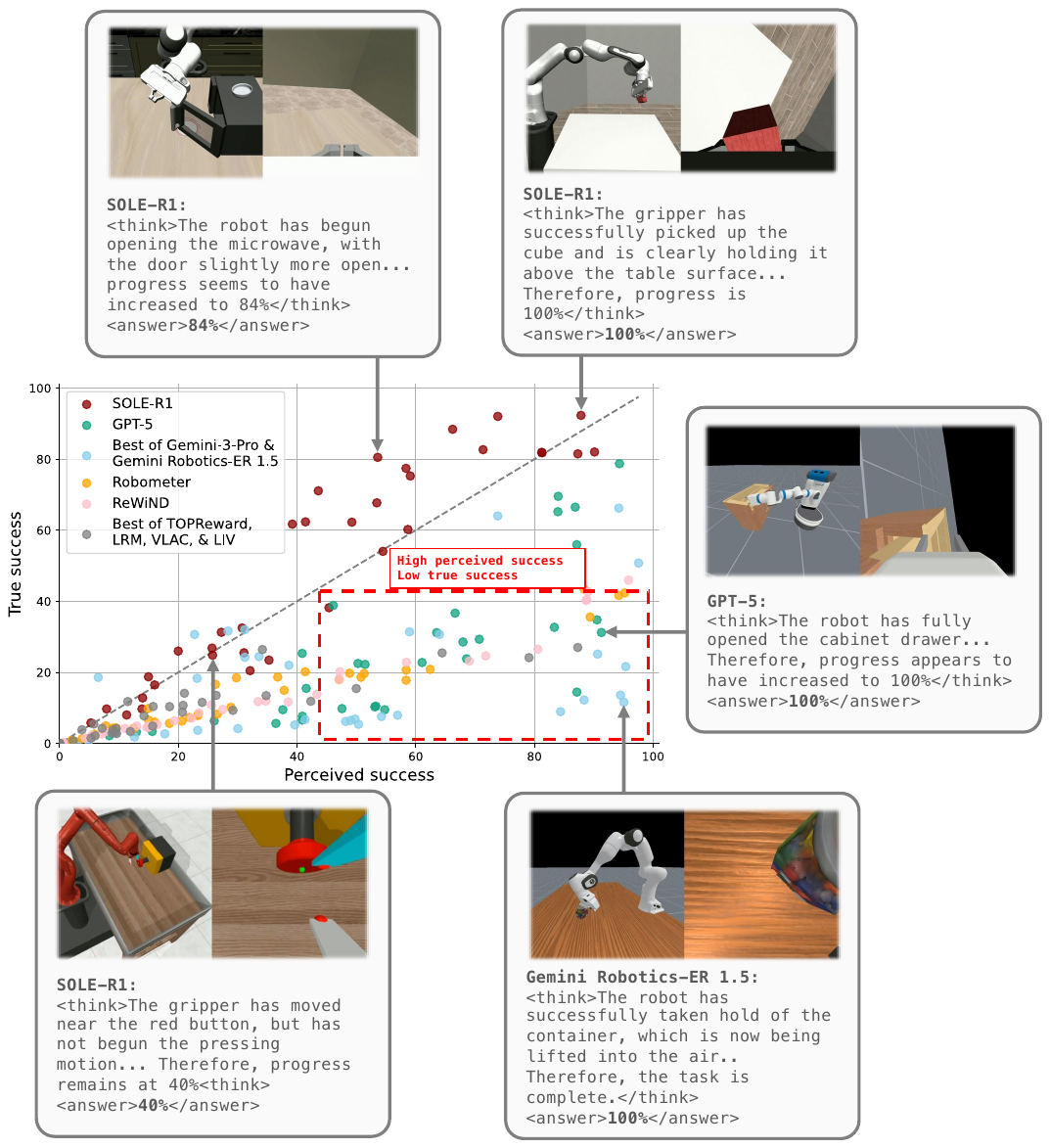}}
      \caption{Extended reasoning examples across all models.}
      \label{fig_reasoning_examples}
  \end{center}
\end{figure}

\newpage
\section{Failure analysis}
\label{sec:failure_analysis}

To understand when and why SOLE-R1 fails to learn a task, we analyze failures at two levels: (i) \emph{task-level reward pathologies} (whether the reward signal is exploitable, miscalibrated, or simply too weak to drive learning), and (ii) \emph{frame-level reasoning errors} (what the model is missing perceptually or temporally when it produces an incorrect progress estimate). Our goal is to distinguish failures due to \emph{reward hacking} from failures due to \emph{insufficient learning signal} under partial observability and distribution shift.

\paragraph{Perceived success vs.\ true success.}
Figure ~\ref{res_fig2} plots, the final policy's \emph{perceived success} based on the predicted rewards (x-axis) versus the \emph{true success} it achieves for each task. This visualization separates failures into two qualitatively different types:

\begin{itemize}
    \item \emph{Reward-hacking failures (high perceived, low true).} Points in the lower-right quadrant indicate that the policy found behaviors that elicit high predicted progress without achieving the task. This is the dominant failure mode for general-purpose VLM rewarders (GPT-5, Gemini-3-Pro, Gemini Robotics-ER), consistent with online RL actively searching for reward loopholes.
    \item \emph{Signal-limited failures (low perceived, low true).} Points in the lower-left quadrant indicate that the learned reward remains pessimistic and the policy never reaches states that the model considers meaningfully closer to the goal. These failures are more common for SOLE-R1 than for baseline VLMs, reflecting that SOLE-R1 is harder to ``trick'' into claiming success, but can still provide a reward that is too noisy or too weak to bootstrap exploration within the episode budget.
\end{itemize}

For SOLE-R1, the majority of failed tasks lie in the \emph{signal-limited} failure type (lower-left quadrant), indicating that the model typically \emph{recognizes} non-success rather than hallucinating completion. This contrasts with GPT-5 and Gemini variants, where many failures appear in the \emph{reward-hacking} failure type (lower-right quadrant), where the learned policy reliably induces high progress predictions while true success remains near zero.

\begin{figure}[ht]
  \begin{center}
    \centerline{\includegraphics[width=.6\columnwidth]{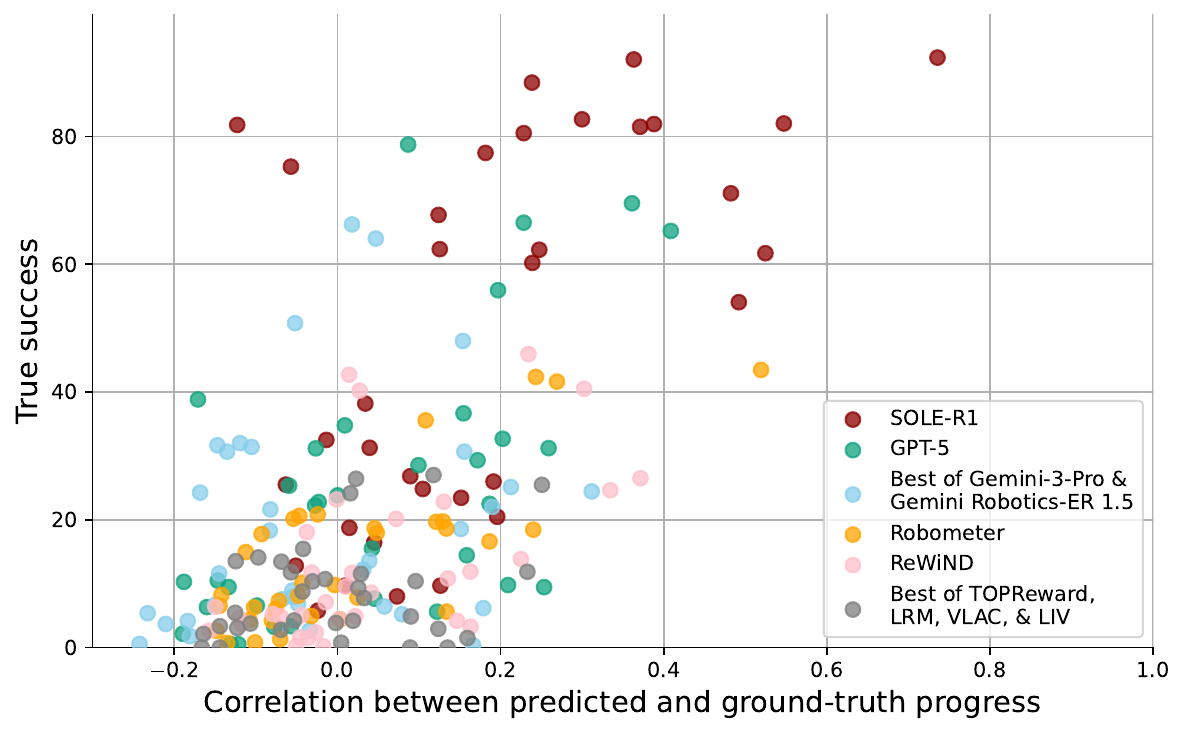}}
    \caption{Zero-shot RL success vs correlation between predicted and ground-truth progress}
    \label{res_fig_corr_vs_true_success}
  \end{center}
\end{figure}

To quantify whether failure is driven by misalignment of the dense reward, we additionally compute the correlation between predicted rewards and ground-truth rewards along trajectories (Figure ~\ref{res_fig_corr_vs_true_success}). We find that for SOLE-R1, many signal-limited failures also exhibit \emph{low reward correlation}, suggesting that the reward does not provide a sufficiently shaped gradient toward success even when the model is not being exploited. In practice, this manifests as the agent oscillating among visually similar non-progress states (e.g., hovering near an object, repeatedly tapping a handle, or regrasping without changing articulation state) with consistently low predicted progress. 

\paragraph{Common failure modes from qualitative review.}
We hand-review 100 rollouts from experiments in simulation and real-world across all methods, and assign each failure to a primary error category (Table~\ref{tab:qual_failure_summary}). We summarize the most common SOLE-R1 failure modes below, and contrast them with baseline VLM rewarders.

\begin{enumerate}
    \item \emph{Temporal under-detection of contact and state transitions.}
    Some tasks hinge on brief events (first contact, latch release, button actuation) that are hard to infer from RGB alone and may occur between model query steps. When these events are missed, progress remains flat despite real advancement, yielding weak learning signal. This occurs most often when (i) the event is visually subtle, (ii) the wrist camera is occluded by the gripper, or (iii) progress depends on a latent state (e.g., a switch toggled) with minimal pixel change.
    
    \item \emph{Ambiguous object state under partial observability.}
    In cluttered scenes, the model can be uncertain whether an object is \emph{actually} grasped, fully inserted, or properly seated, especially when only one viewpoint clearly resolves the state. SOLE-R1 tends to respond conservatively in these cases (low $p_t$), which reduces reward hacking but can prevent the agent from receiving positive reinforcement for near-miss states that are essential stepping stones.

    \item \emph{Over-reliance on goal-consistent appearance cues.}
    Rarely, SOLE-R1 assigns partial progress based on visual cues that correlate with success (e.g., the gripper near a handle, an object aligned with a receptacle) even when the underlying subgoal is not achieved (handle not pulled, object not released). Unlike baseline VLMs, these errors typically saturate at \emph{moderate} progress and do not produce confident ``100\%'' completion, but they can still bias exploration toward non-productive states.

\end{enumerate}

In contrast, baseline VLM rewarders fail predominantly via \textbf{perceptual hallucination}: they often infer success even when observations contradict it (e.g., interpreting occlusion as lifting, interpreting proximity as grasp). This directly enables reward hacking, where the policy learns to place the camera or objects into configurations that elicit confident success statements without completing the task (Figure~\ref{res_fig2} and \ref{fig_reasoning_examples}).

\begin{table}[t]
\centering
\footnotesize
\caption{
Quantitative summary of failure modes from qualitative review of real-robot and simulation rollouts.
}
\begin{tabular}{lcccc}
\toprule
\textbf{Primary qualitative error mode} 
& \textbf{SOLE-R1} 
& \textbf{GPT-5} 
& \textbf{Gemini-3-Pro} 
& \textbf{Gemini Robotics-ER} \\
\midrule
Temporal under-detection of contact / state change 
& 34\% & 14\% & 17\% & 19\% \\
Ambiguous object state under partial observability 
& 29\% & 11\% & 13\% & 15\% \\
Relies on goal-consistent appearance cues 
& 15\% & 18\% & 20\% & 21\% \\
Perceptual hallucination of success 
& 9\% & 42\% & 38\% & 35\% \\
Other / unclassified 
& 13\% & 15\% & 12\% & 10\% \\
\bottomrule
\end{tabular}
\label{tab:qual_failure_summary}
\end{table}

\section{Ablations}
\label{sec:ablations}

We ablate SOLE-R1 to isolate components responsible for reasoning that (i) is robust to reward hacking and (ii) provides usable learning signal for unseen tasks. We evaluate three ablations that remove (A) \text{explicit natural language CoT reasoning} during training, (B) authentic \text{negative / non-expert trajectories} from robot video progress prediction training, and (C) \text{foundational spatiotemporal reasoning data} used to induce general spatial and multi-frame temporal grounding.

\paragraph{Evaluation protocol.}
Unless otherwise stated, all ablations share the same backbone VLM, training protocol, compute budget, and online RL algorithm. We re-run the full zero-shot online RL benchmark suite from Section~\ref{sec:zero_shot_rl} with identical hyperparameters and compute (i) task success rate, (ii) perceived-vs-true success scatter (Figure \ref{res_fig3}), and (iii) reward/ground-truth alignment metrics. We also report the fraction of failures categorized as \emph{reward-hacking} (high perceived / low true) versus\ \emph{signal-limited} (low perceived / low true) using the same quadrant thresholds as Section~\ref{sec:failure_analysis_compressed}.

\paragraph{Ablation 1: No CoT reasoning in progress training.}
We remove the \texttt{<think>} channel from progress training. The model is trained to directly emit only \texttt{<answer>} (the scalar progress) at each timestep, without generating per-timestep natural-language explanations:
\[
y_t = [\texttt{<answer>}~p_t~\texttt{</answer>}].
\]
Generating temporally grounded rationales acts as an auxiliary task that forces the model to attend to \text{state changes} (contacts, releases, articulation transitions) rather than shortcut correlations. This should improve calibration and reduce variance in dense progress estimates, especially under occlusion or viewpoint shift.

\textbf{Results.} Removing CoT consistently degrades learning across tasks and shifts performance toward the \emph{signal-limited} failure type (Figure \ref{res_fig3}): the reward becomes flatter and noisier, yielding weaker incremental credit assignment. Qualitatively, thinking-ablated models often under-react to small but task-critical transitions (e.g., closing fingers to grasp an object), producing progress curves with long plateaus and delayed spikes.

\paragraph{Ablation 2: Expert-only robot-video progress supervision.}
We remove authentic unsuccessful / non-expert trajectories from the robot-video progress dataset, training only on near-expert rollouts. However, we keep the artificial non-expert trajectories, generated by randomly reversing the video at random moments. This augmentation prevents the model from learning to predict monotonically increasing progress values, but is limited in the diversity of failure states it can mimic.

In online RL, the agent frequently visits off-distribution states and executes partial, incorrect, or exploitative behaviors. Training only on successful trajectories can lead to overly optimistic extrapolation: states that are merely \emph{goal-adjacent} (gripper near handle, object aligned with receptacle) may be scored as progress even when no causal subgoal is achieved. We hypothesize that authentic negative and recovery trajectories (generated by inserting real random actions within expert trajectories as described in Section~\ref{sec:progress_prediction_data}) are critical for familiarizing the model with real failure states and helping it learn \text{counterfactual} distinctions (``looks close'' vs.\ ``is achieved'') in a way that supports robustness to distribution shift.

\textbf{Results.} Expert-only training substantially increases the \emph{reward-hacking} failure types (Figure \ref{res_fig3}). The learned reward becomes easier to exploit: agents discover configurations that mimic visual correlates of success (e.g., hovering near target objects, aligning poses, occluding failure) that trigger inflated progress predictions. On the perceived-vs-true plot, these failures move from the lower-left quadrant (signal-limited) toward the lower-right quadrant (reward hacking), indicating higher predicted progress without corresponding true success. This mirrors the baseline VLM rewarder pathology, though typically less extreme than GPT-5/Gemini.

\paragraph{Ablation 3: Remove general spatial \& multi-frame reasoning data.}
We remove the general spatial reasoning and multi-image temporal reasoning datasets used in the foundational stage (Section~\ref{sec:training_data}), and train only on embodied reasoning and planning, along with robot-video progress estimation data.

Robot progress estimation requires recognizing subtle spatial relations (contact, containment, alignment, occlusion) and temporally local changes. Without broad spatial + multi-frame reasoning pretraining, the model may rely on dataset-specific appearance cues and fail under new scenes, viewpoints, embodiments, and task families. We hypothesize that foundational reasoning data improves generalization and reduces reliance on spurious correlations that online RL can exploit.

\textbf{Results.} Removing foundational reasoning data reduces overall success and again increases \emph{reward-hacking} failure types. This abaltion produces a model that is more sensitive to camera viewpoint and scene texture changes: the agent can more easily induce states that resemble training-time success patterns while failing at the underlying task (e.g., partial occlusion producing ``hallucinated'' grasp).

\paragraph{Summary.}
Removing CoT primarily harms \emph{signal quality} (more signal-limited failures), while removing authentic negative trajectories or foundational reasoning data primarily harms \emph{robustness} (more reward-hacking failures). These results support our central design: (i) temporally grounded language reasoning is an effective scaffold for dense progress prediction, and (ii) broad spatiotemporal grounding plus hard negatives are essential for resisting exploitation in zero-shot online RL.

\section{Effect of RLVR beyond SFT}
\label{appendix:ablation_sft_vs_rlvr}

\begin{figure}[H]
  \begin{center}
    \centerline{\includegraphics[width=.6\columnwidth]{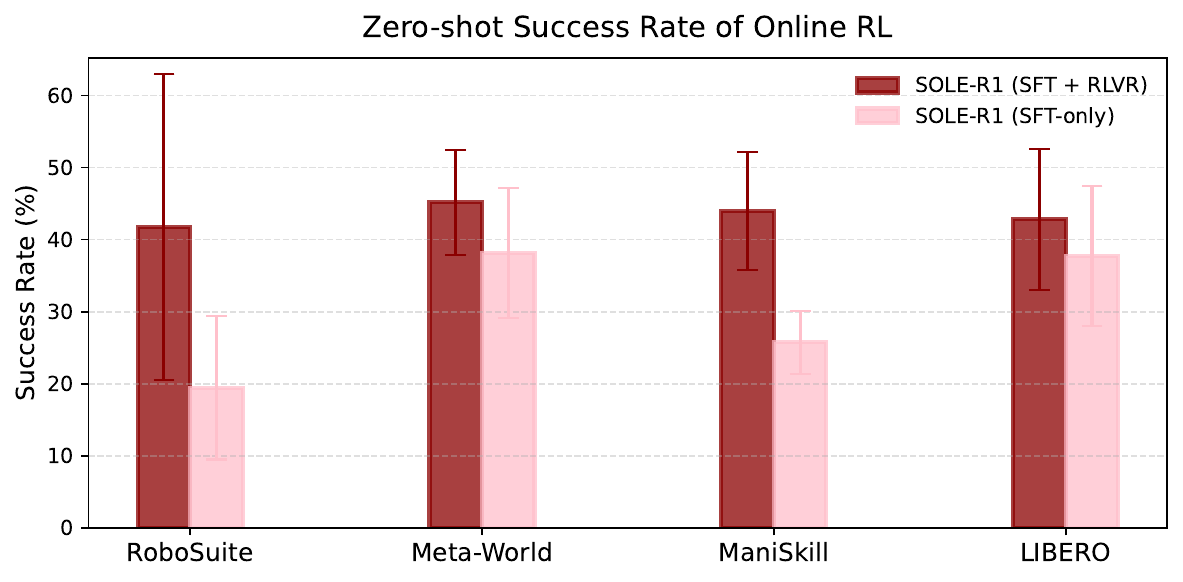}}
    \caption{SFT-only vs SFT+RLVR model in zero-shot success rate of online RL}
    \label{res_fig_sft_only}
  \end{center}
\end{figure}

We isolate the contribution of reinforcement learning with verifiable rewards (RLVR) by comparing a model trained with \textbf{SFT only} to the full \textbf{SFT+RLVR} SOLE-R1 model. Both models share the same backbone, training data, prompts, and inference procedure; the only difference is the additional RLVR stage applied to the latter. Figure~\ref{res_fig_sft_only} reports zero-shot online RL success rates for both variants.

\paragraph{Results.}
Across environments, adding RLVR yields a clear and consistent improvement in zero-shot RL performance. The SFT-only model is able to provide a usable reward signal on many tasks, indicating that supervised spatiotemporal CoT reasoning alone already induces partial alignment between predicted progress and true task advancement. However, its performance is less reliable than the full model. In contrast, the SFT+RLVR model achieves consistently higher success rates, solving more tasks and reaching higher peak performance on tasks that are partially solvable with SFT alone.
This gap highlights a key limitation of pure SFT for reward modeling: during supervised training, the scalar progress value in the \texttt{<answer>} channel constitutes only a small fraction of the output tokens, and its numerical accuracy is weakly emphasized relative to generating fluent reasoning text. 
RLVR directly addresses this limitation by concentrating learning signal on the correctness and calibration of the predicted progress value. 
Importantly, RLVR operates on top of the strong spatiotemporal reasoning induced by SFT, refining \emph{how much} progress is assigned without degrading the underlying reasoning structure.


\section{Full related work}
\label{appendix:full_related_work}

\paragraph{On-robot RL}

Applying reinforcement learning directly on physical robots has long been challenging due to sample inefficiency, safety concerns, and the difficulty of specifying robust reward functions. Seminal works in this area rely on carefully engineered rewards and resets in the real world \cite{Riedmiller2009RobotSoccer,Levine2016EndToEndVisuomotor}, massively parallelized actors \cite{Levine2018HandEyeCoordination} and asynchronous learning algorithms \cite{gu2017deep}, spatial inductive biases encoded into neural networks \cite{zeng18learning,wang22onrobot} and residual learning on top of planner trajectories \cite{zeng2020tossingbot}, and reward learning by human-in-the-loop labeling \cite{singh2019}.

Recent on-robot learning works increase sample efficiency and the complexity of solvable tasks by highly tuned and efficient off-policy learning implementations \cite{Luo2024SERL}, human-in-the-loop learning \cite{Luo2025bDexterousHIL}, behavior priors in mobile manipulation \cite{Mendonca2024AutonomousMobileManipRL} and shaped reward function learning \cite{Yang2024aRank2Reward,biza2025robot}. In parallel, the emergence of generalist robot policies trained on large-scale datasets \cite{brohan2022rt1,OctoModelTeam2024Octo,Kim2024OpenVLA,Black2024Pi0} has shifted attention toward methods that fine-tune, refine or steer such policies using RL during deployment \cite{Zhang2024GRAPE,Mark2024PolicyAgnosticRL,Nakamoto2024SteeringGeneralists,Chen2025bConRFT,Hu2025FLARE,Ankile2025ResidualRLAssembly,Wagenmaker2025SteerDiffusionPolicy,ankile2025residual,lei2025rl}. 

Despite these advances, nearly all existing approaches still rely on either manually specified reward functions or human-provided supervision. In contrast, SOLE-R1 targets a setting where \emph{no ground-truth reward, demonstrations, or task-specific tuning} are available, and learning must be driven entirely by a pretrained video-language reasoning model.



\paragraph{Vision-language reward learning}

With the rise of vision-language models, recent work has explored using VLMs as sources of reward supervision. Preference-based methods query VLMs for pairwise comparisons or ratings to learn reward functions \cite{Wang2024RLVLMF,Venkataraman2024RealWorldOfflineRLFromVLMFeedback,Luu2025ERLVLM,Singh2025VARP}. Other approaches directly score individual images or videos to produce sparse or dense rewards \cite{Du2023SuccessDetectors,Rocamonde2023ZeroShotRewardModels,Baumli2023VLMRewards,Yang2024aRank2Reward,Alakuijala2024VideoLanguageCritic,Yang2024bAdapt2Reward,Venuto2024CodeAsReward}. 


Several recent methods are especially relevant to our setting. RoboReward \cite{lee2026roboreward}, Robometer \cite{liang2026robometer}, and LRM \cite{wu2026large} train vision-language models to predict task progress from robot videos. However, these models typically (i) do not produce explicit intermediate reasoning, (ii) are trained primarily on expert or near-expert trajectories, and (iii) are not evaluated as the \emph{sole} reward signal for online RL starting from random policies. Our experiments show that these limitations make them vulnerable to reward hacking and miscalibration under distribution shift.

\paragraph{Temporal supervision and counterfactual trajectories.}
Several works exploit temporal structure in videos to derive supervision without explicit rewards. Using temporal order as a proxy for progress has been explored for value learning and reward relabeling \cite{Ma2024InContextValueLearners,Zhang2025aRewind}, and related ideas appear in non-robot domains such as video understanding and representation learning. However, prior methods generally ignore the semantic content of the trajectory or treat time as a weak ordinal signal. 

In contrast, SOLE-R1 combines temporal-order supervision with explicit spatiotemporal reasoning grounded in video evidence. By training on both authentic non-expert trajectories and counterfactual regressions, our model learns to distinguish genuine causal progress from visually goal-adjacent but incorrect states, which is critical for robustness in online RL.

\paragraph{Process-level and reasoning-based reward models.}
Outside robotics, reinforcement learning from learned reward models has been extensively studied for post-training large language models \cite{Lightman2023LetsVerifyStepByStep,Shao2024DeepSeekMath,guo2025deepseek,Luo2025aWizardMath,luounlocking}. These works highlight the importance of process-level supervision and verifiable rewards for stabilizing learning.
Within robotics, Tan et al.~\cite{Tan2025RoboDopamine} propose a concurrent process-reward modeling approach for high-precision manipulation. While complementary, their work focuses on narrow task families and does not study zero-shot online RL across diverse robots and environments. SOLE-R1 instead emphasizes \emph{general-purpose video-language reasoning} that produces dense progress estimates applicable across tasks, embodiments, and viewpoints, and directly evaluates whether such reasoning can serve as the sole learning signal for RL.

\section{Reinforcement learning agent details}
\label{sec:rl_hyperparameters}

Our RL agent is a PyTorch re-implementation of a DrQv2 \cite{yarats22mastering} agent from SERL \cite{Luo2024SERL} (JAX). For tasks in Meta-World and real world, we adapt a Soft Actor-Critic implementation from Tianshou \cite{weng22tianshou} to match the SERL implementation. For tasks in  RoboSuite, LIBERO, and ManiSkill, we implement a DrQv2 agent that builds upon prior work from \cite{dalal2024plan} that previously showed success for tasks based in the RoboSuite simulation environment. We train the actor and critic from scratch, without any pre-training or replay buffer demonstrations. Our hyper-parameters are listed in Table \ref{tab:hyperparameters}.

\begin{table}[H]
    \centering
    \footnotesize
    \caption{Reinforcement learning agent details}
    \begin{tabular}{cccc}
        \toprule
        DrQv2 & Meta-World & RoboSuite, LIBERO, ManiSkill & Real-world \\
        \midrule
        Buffer size & 200000 & 200000 & 200000 \\
        Batch size & 256 & 256 & 256 \\
        Critic to actor steps & 4 & 4 & 4 \\
        Initial random steps & 1000 & 2500 & 100 \\
        Start learning at & 1000 & 2500 & 100 \\
        Critic ensemble size & 10 & 10 & 10 \\
        Critic subsample size & 2 & 2 & 2 \\
        Discount & 0.96 & 0.99 & 0.9 \\
        Exploration schedule & Entropy tuning with $\alpha=10^{-2}$ & std=linear(1.0, 0.1, 500K) & std=linear(1.0, 0.1, 15k) \\
        Learning rate & $10^{-3}$ & $10^{-4}$ & $10^{-3}$ \\
        Soft target update rate & $5 * 10^{-3}$ & $5 * 10^{-3}$ & $5 * 10^{-3}$ \\
        Learner steps per actor step & 1 & 2 & 4 \\
        \bottomrule
    \end{tabular}
    \label{tab:hyperparameters}
\end{table}

\begin{table}[H]
\caption{Task groups and their natural language specifications}
\centering
\footnotesize
\begin{tabular}{ll}
\hline
\textbf{Task Name} & \textbf{Natural Language Task Specification} \\
\hline
\multicolumn{2}{l}{\textbf{Real-World Tasks}} \\
\hline
Pick can & Pick up the can \\
Touch strawberry in clutter & Touch the strawberry on the table \\
Push can to cube & Push the can near the cube \\
Close drawer & Close the drawer \\
Open drawer & Open the drawer \\
Insert pipe & Insert the pipe into the opening \\
\hline
\multicolumn{2}{l}{\textbf{Meta-World Tasks}} \\
\hline
faucet-open & Open the faucet \\
faucet-close & Close the faucet \\
window-open & Open the window \\
window-close & Close the window \\
plate-slide (puck in net) & Slide the puck into the net \\
drawer-close & Close the drawer \\
drawer-open & Open the drawer \\
door-close & Close the door \\
door-open & Open the door \\
button-press & Press the button \\
handle-pull & Pull the handle \\
soccer & Push the soccer ball into the goal \\
lever-pull & Pull the lever up \\
coffee-button & Press the gray coffee machine button \\
handle-press & Press the handle down \\
coffee-push & Push the coffee mug toward the coffee machine \\
\hline
\multicolumn{2}{l}{\textbf{RoboSuite Tasks}} \\
\hline
Lift & Pick up the cube from the table \\
Wipe & Wipe the dark ink off of the table \\
Door & Turn the door handle and pull the door open \\
PickPlaceCan & Pick up the can and place it at a target location \\
\hline
\multicolumn{2}{l}{\textbf{LIBERO Tasks (Environment Task ID)}} \\
\hline
Close microwave {\footnotesize(\text{KITCHEN\_SCENE6\_close\_the\_microwave})} & Close the microwave door \\
Open microwave {\footnotesize(\text{KITCHEN\_SCENE7\_open\_the\_microwave})} & Open the microwave door \\
Turn on stove {\footnotesize(\text{KITCHEN\_SCENE3\_turn\_on\_the\_stove})} & Turn the stove knob to the on position \\
Turn off stove {\footnotesize(\text{KITCHEN\_SCENE8\_turn\_off\_the\_stove})} & Turn the stove knob to the off position \\
Close top drawer {\footnotesize(\text{SCENE10\_close\_the\_top\_drawer\_of\_the\_cabinet})} & Close the top drawer \\
Open bottom drawer {\footnotesize(\text{SCENE1\_open\_the\_bottom\_drawer\_of\_the\_cabinet})} & Open the bottom drawer \\

\hline
\multicolumn{2}{l}{\textbf{ManiSkill Tasks (Environment Task ID)}} \\
\hline
PickPanda (PickCube-v1) & Pick up the cube from the table \\
PushCube (PushCube-v1) & Push the cube into the target area \\
PullCube (PullCube-v1) & Pull the cube into the target area \\
Mobile-OpenCabDrawer (OpenCabinetDrawer-v1) & Open the cabinet drawer\\
PickSO100 (PickCubeSO100-v1) & Pick up the cube from the table \\
PickWidowXAI (PickCubeWidowXAI-v1) & Pick up the cube from the table \\
PickSingleYCB  (PickSingleYCB-v1) & Pick up the object from the table \\
\hline
\end{tabular}
\end{table}

\subsection{Evaluation protocol and success rate computation}
\label{sec:success_rate_eval}

We evaluate the learned policies using a task-level \emph{success rate} that is computed from ground-truth environment signals and task-specific termination conditions, and is \emph{never} observed by the agent during training.

\paragraph{Evaluation episodes.}
For each task, we periodically evaluate the current policy by running $N$ independent episodes (typically $N=20$ in simulation and $N=10$ in real-world experiments). Each episode is initialized from the standard task reset distribution and is executed for a fixed horizon of $H$ steps. During evaluation, exploration noise is disabled and the policy acts deterministically.

\paragraph{Success definition.}
A rollout is considered \emph{successful} if the task-specific ground-truth success condition is satisfied at \emph{any} timestep during the episode. These conditions are provided by the simulator (for RoboSuite, ManiSkill, Meta-World, and LIBERO) or by manually defined, externally measured criteria in the real world (e.g., drawer closed within a tolerance, object lifted off the table, object reaching a target region). Importantly, these success signals are used \emph{only} for evaluation and logging, and are not accessible to the policy or the reward model during training.

\paragraph{Binary episode outcome.}
Let $s_i \in \{0,1\}$ denote the binary success outcome of evaluation episode $i$, where $s_i = 1$ if the success condition is met at any timestep and $s_i = 0$ otherwise. We do not require the success condition to persist until the end of the episode; transient successes (e.g., briefly lifting an object) are counted as success, consistent with prior work in online manipulation RL.

\paragraph{Success rate.}
The success rate for a task is computed as
\[
\text{SuccessRate} = \frac{1}{N} \sum_{i=1}^{N} s_i.
\]
For each reported result, we repeat training with multiple random seeds (three seeds in simulation and one seed in real-world experiments) and report the mean success rate across seeds. Error bars in plots denote the standard error across seeds.



\section{Vision-language model inference details}
\label{sec:vlm_inference}

\subsection{Inference speed}

We send a batch of 2 videos of 20 frames each to SOLE-R1, which is hosted on a node with a single H100, and to GPT-5 and Gemini-3-Pro using public APIs. We set the Gemini-3-Pro thinking setting to ``LOW''. On average, SOLE-R1 takes 40 seconds to annotate rewards (1 s per frame), GPT-5 takes 366 seconds (9.15 s per frame) and Gemini-3-Pro takes 322 seconds (8.05 s per frame). All three models produce a chain of thought for each frame, and the inference speed may vary according to the output length each model chooses to produce. We note that the speed of the public APIs varies day by day.

\subsection{Prompts}

Each model receives a system prompt followed by a user prompt. A sequence images follows. In user prompt, we replace ``\{task\_description\}'' with the task description and ``\{prev\_progress\}\'' with the previous model prediction. Since we give the model its previous prediction, we have to perform inference sequentially.

\subsubsection{SOLE-R1}

SYSTEM PROMPT TEMPLATE:
You are an expert roboticist with the goal of predicting task progress percentages given frames from a video of a robot attempting to complete a task. You first think, in the form of an internal monologue, before providing your final answer. Your reasoning process MUST BE enclosed within $<$think$>$ $<$/think$>$ tags and should include detailed reasoning. Your final answer MUST BE enclosed within $<$answer$>$ $<$/answer$>$ tags and should be a integer (positive or negative) representing current task progress percentage. ``Example output format: $<$think$>$[detailed reasoning process]$<$/think$>$$<$answer$>$[current task progress]\%$<$/answer$>$

USER QUESTION TEMPLATE:
Here is an image containing multiple camera views of a robot attempting to complete a task. The views on the top are from an external camera. The views on the bottom are from the robot's wrist camera. The views from the very first timestep are shown to the left. The views from the previous timestep are shown in the middle. The views from the current
timestep are shown to the right. The task description is: \{task\_description\}. The task progress for the very first timestep is 0\%. The task progress for the previous timestep is \{prev\_progress\}\%. Predict the task progress for the current timestep.

\subsubsection{GPT-5}

SYSTEM PROMPT: None.

USER PROMPT TEMPLATE GPT:
Here is an image containing multiple camera views of a robot attempting to complete a task. The first image is from the previous timestep. The second image is from the current timestep. The task description is: \{task\_description\}. The predicted task progress for the previous timestep was \{prev\_progress\}\%. Predict the task progress for the current timestep. Before providing your final answer, first briefly provide a few sentences that reason about what is happening at the current timestep relative to the previous timestep. Your reasoning process should be no more than a few sentences and MUST BE enclosed within $<$think$>$ $<$/think$>$ tags. Please refer to the images as timesteps instead of as separate images. Your final answer MUST BE enclosed within $<$answer$>$ $<$/answer$>$ tags and should be an integer representing current task progress percentage. Example output format: $<$think$>$[detailed reasoning process]$<$/think$>$$<$answer$>$[current task progress]\%$<$/answer$>$

\subsubsection{Gemini-3-Pro and Geminin Robotics ER 1.5}

SYSTEM PROMPT:
You are an expert roboticist with the goal of predicting task progress percentages given frames from a video of a robot attempting to complete a task.

USER PROMPT TEMPLATE GEMINI:
Here is an image containing multiple camera views of a robot attempting to complete a task.  The first image is from the previous timestep. The second image is from the current timestep. The task description is: \{task\_description\}. The predicted task progress for the previous timestep was \{prev\_progress\}\%. Predict the task progress for the current timestep. Note that the previous progress value might not be accurate. Please carefully assess the images to determine the correct progress for the current timestep. Also note that the performance of the robot is unknown so progress can increase or decrease at any timestep. Before providing your final answer, first briefly provide a few words that reason about what is happening at the current timestep relative to the previous timestep. Your reasoning process should be no more than one or two sentences and MUST BE enclosed within $<$think$>$ $<$/think$>$ tags. IMPORTANT: Do not refer to 'the user' in your reasoning. Only reason as though it is an internal monologue thinking about the robot. If you are unsure about the progress, please try to avoid overestimation and provide a conservative estimate. Please refer to the images as timesteps instead of as separate images. Your final answer MUST BE enclosed within $<$answer$>$ $<$/answer$>$ tags and should be an integer representing current task progress percentage. Example output format: <think>[detailed reasoning process]$<$/think$>$$<$answer$>$[current task progress]\%$<$/answer$>$

\subsubsection{Testing different prompting strategies for GPT-5 and Gemini}
\label{appendix:prompting}
\begin{figure}[H]
  \begin{center}
    \centerline{\includegraphics[width=.6\columnwidth]{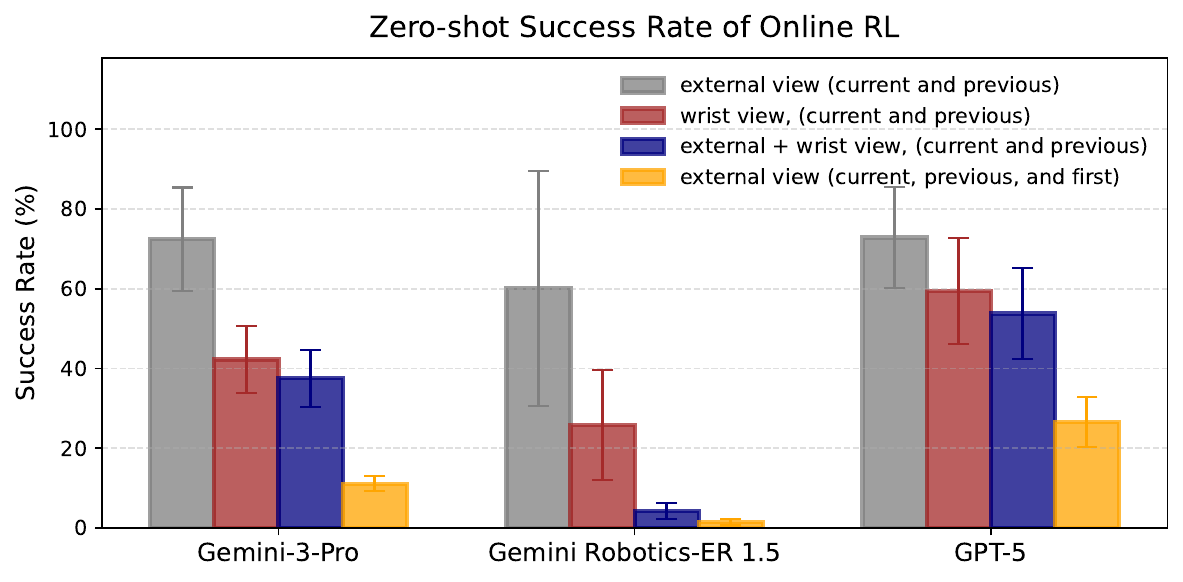}}
    \caption{Zero-shot success rate of online RL on 10 simple Meta-World tasks across varying prompting strategies.}
    \label{res_fig_prompting}
  \end{center}
\end{figure}

Figure~\ref{res_fig_prompting} compares several prompting strategies for GPT-5 and Gemini in zero-shot online RL, varying the camera viewpoints and temporal context provided to the model.
We find that prompting with the \emph{external camera view only}, using the \emph{current and previous} timesteps, consistently yields higher task success than alternatives that include (i) wrist-view images, (ii) a combination of external and wrist views, or (iii) longer temporal context including the first timestep.
This result suggests that, for general-purpose VLM rewarders, external views provide more reliable and less ambiguous cues for task progress than wrist-mounted views, which are frequently occluded or dominated by the robot end-effector.
Additionally, providing longer temporal histories does not improve performance and can degrade it, likely due to increased visual clutter and reduced salience of task-relevant changes.
All baseline results in the main paper therefore use the external-view (current + previous) prompting configuration.

\section{Real robot experiment details}
\label{sec:real_world_details}

\begin{figure}[H]
    \centering
    \begin{subfigure}{.721\textwidth}
        \includegraphics[width=\linewidth]{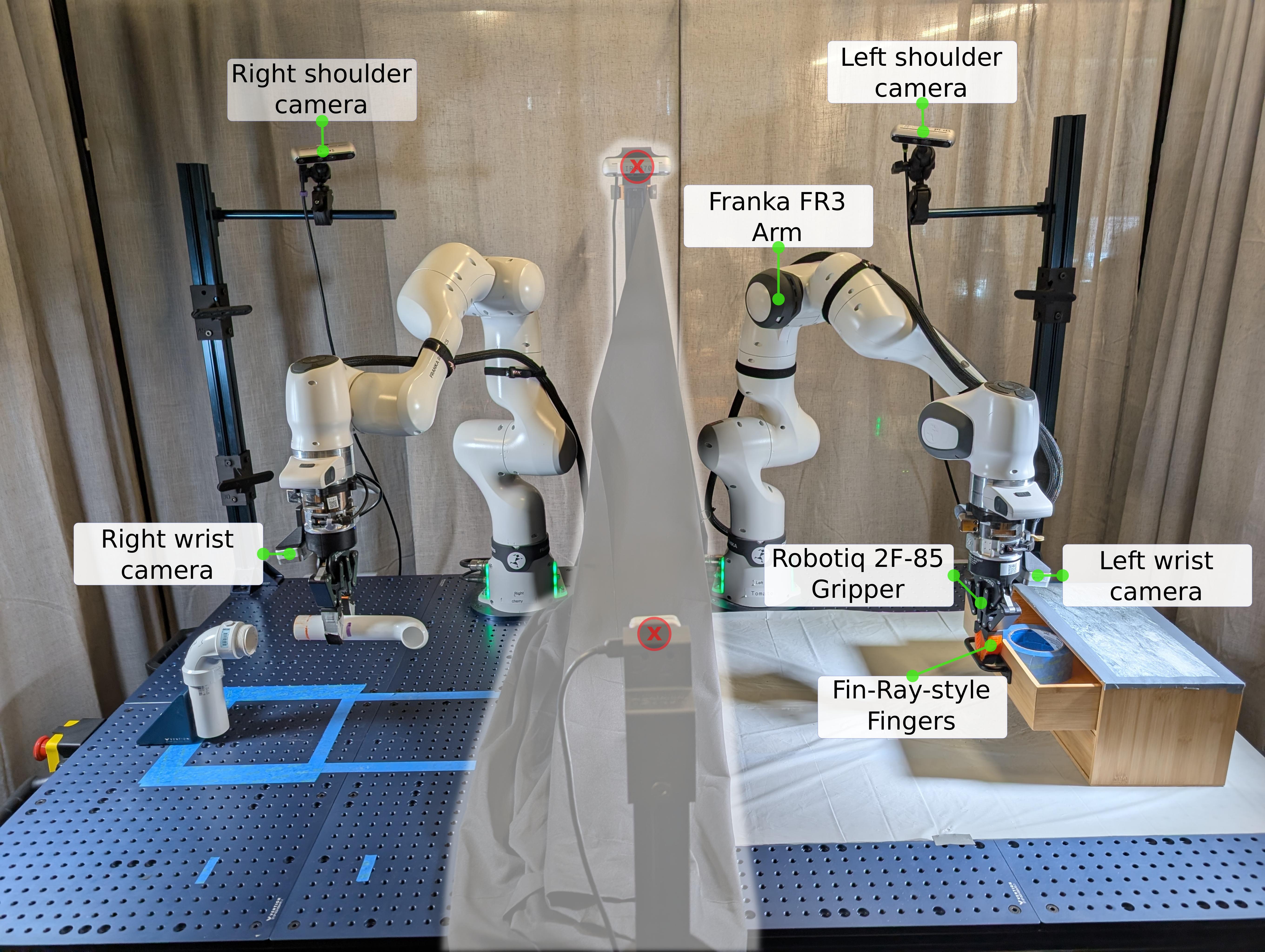}    
    \end{subfigure}%
    \vspace{0.1\textwidth}
    \begin{subfigure}{.269\textwidth}
        \includegraphics[width=\linewidth]{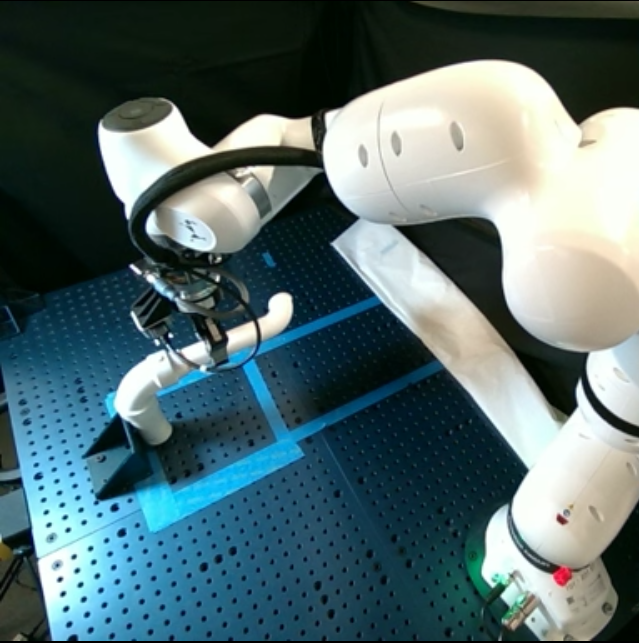} \\
        \includegraphics[width=\linewidth]{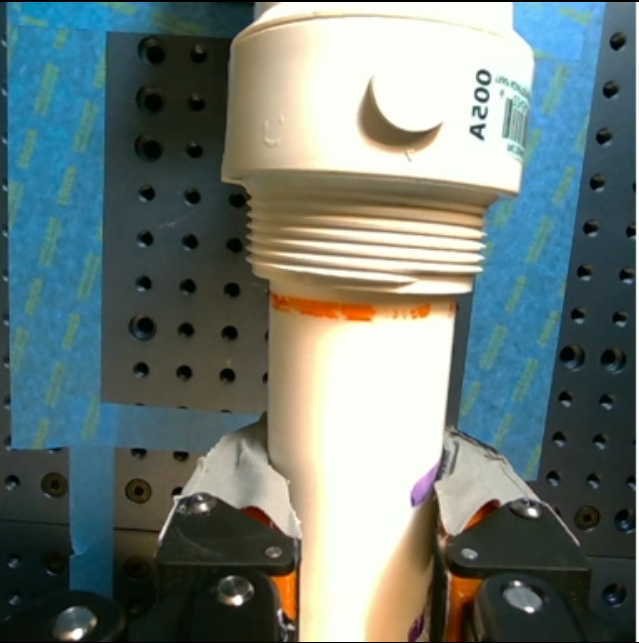}
    \end{subfigure}
    \caption{Left: workcell with two Franka FR3 arms. Right: example shoulder and wrist views from the right robot.}
    \label{fig:real_robot_station}
\end{figure}

We conduct visuomotor reinforcement learning (RL) experiments on real tabletop Franka FR3 arms. Our system consists of an RL actor, which executes actions on the robot, an RL learner, which performs gradient descent based on collected data, and a reward labeling server, which provides reward annotations for collected episodes. The RL actor, RL learner and reward labeling server run as independent processes and communicate asynchronously via \href{https://zeromq.org/}{ZeroMQ}. On average, our system executes around 1.5k RL actions and 6k RL learner steps per hour. We reset the environment manually; the arms are reset using either an automatic motion plan or manually via SpaceMouse teleoperation. There is no extrinsic reward signal, the only supervision comes from our reward labeling server, which uses either our SOLE-R1 vision-language model (VLM) or other baseline VLMs. We either self-host VLMs on nodes equipped with a single H100 GPU or use public APIs.

\paragraph{Hardware configuration} We conduct our experiments on workstations with a pair of Franka FR3 arms mounted on a shared tabletop (Figure \ref{fig:real_robot_station}). The Franka end-effectors are equipped with a wrist-mounted RealSense D405 camera, a force-torque sensor, which is not used in our experiments, and a Robotiq 2F-85 gripper with Fin-Ray-style compliant fingers. While there are multiple cameras mounted to the workstation, we only use a pair of cameras looking over the shoulder of each of the robots and the pair of wrist-mounted cameras. We run RL experiments independently on each arm.

\paragraph{Observation space} Our observation space combines visual inputs and proprioception. We record RGB images from a wrist-mounted camera and a over-the-shoulder camera. The wrist cameras are RealSense D405s and the shoulder cameras are D455s. We crop and rescale these images to (384, 384) for reward prediction and to (84, 84) for action prediction. We add barriers so that only one robot is seen at a time. We add proprioceptive state in the form of end-effector position (x, y, z) and gripper width.

\paragraph{Action space} The RL actor commands Cartesian end-effector position offsets in (x, y, z) coordinates in the interval [-1, 1]. We internally rescale these actions to [-20 cm, +20 cm] for each axis. We set a fixed gripper orientation for each task and do not allow the RL actor to change it (conversely, we use the full $SE(3)$ action space in simulation). Depending on the task, the RL actor can also command gripper actions scaled between [-1, 1]. We threshold these actions as fully open [1, 0] and fully closed (0, -1]. While we have the ability to continuously command the gripper width, this mode is not conducive to exploration. On average, the RL actor half-closes the gripper, which often prevents it from grasping objects. We therefore use binary gripper control.

\paragraph{Control stack} Our implementation combines an RL control loop and an internal robot control loop. Firstly, the RL actor interacts with the robot environment at 1 Hz: we execute an action, wait 1 s and record the next observation. Secondly, the internal robot control loop executes RL actions at 100 Hz using linear interpolation for positions and spherical linear interpolation (slerp) for quaternion orientations. Each RL action is interpolated into a 1 s trajectory with 100 points. Then, we perform \href{https://github.com/stephane-caron/pink}{\texttt{pink}} inverse kinematics and execute the resulting joint-space actions using a hybrid joint impedance controller implemented in \href{https://github.com/ros-controls/ros2_control}{\texttt{ros2\_control}}. A reference implementation of this controller can be found in \href{https://github.com/facebookresearch/fairo/blob/main/polymetis/polymetis/python/torchcontrol/policies/impedance.py}{\texttt{polymetis}}.

\paragraph{Reward Prediction} For real-world tasks, our RL episodes are 20 timesteps long. SOLE-R1 labels all frames whereas GPT-5 and Gemini-3-Pro subsample to 10 frames. Even so, these baselines take more than twice as long to label the videos, resulting in experiment duration of up to 5 hours. We end SOLE-R1 experiments after 2 hours. SOLE-R1 receives a composite frame of the shoulder and wrist view, whereas GPT-5 and Gemini-3-Pro only use the shoulder view. We made this choice because both baselines struggle to understand wrist and composite views (Section \ref{appendix:prompting}). In particular, they sometimes state that there are multiple robots in the scene based on the composite view.

\subsection{Tasks}

\paragraph{Touch strawberry in clutter} We place a box, a can and a strawberry randomly in a $20{\times}20{\times}20$ cm workspace. The workspace is slightly higher than the ground to prevent the gripper from scraping the table. The robot starts in a fixed pose above the workspace. The task is successful if the gripper touches the strawberry at any point in the episode.

\paragraph{Push can to cube} We use the same workspace as in the first task. We place a can on the left side of the workspace and a blue cube on the right side of the workspace. Both object positions vary with small random offsets. The robot starts in a fixed pose above the workspace. The task is successful if the can ends up within 1 cm of the cube.

\paragraph{Pick can} We use the same workspace as in the first task. The can is placed approximately in the center with small random offsets. The robot starts in a fixed pose above the workspace. The task is successful if the robot visibly lifts the can off the table at any point in the episode. 

\paragraph{Close drawer} We use a $10{\times}20{\times}5$ cm workspace, we use a small height variation to protect the wrist camera. The cabinet is placed in approximately the same position on the right side of the workspace and the drawer is open up to a point marked by tape. The agent is successful if it closes the drawer up to about the last 1 cm, marked by tape, at any point in the episode.

\paragraph{Open drawer} We use the same workspace as in ``close drawer''. The drawer starts fully closed. The task is successful if the agent opens the drawer up to a point about 10 cm inside of the drawer marked by tape at any point in the episode.

\section{Training data synthesis and curation details}
\label{appendix:data_synth}

This appendix provides additional details on the construction, balancing, and curation of the training data used for SOLE-R1, expanding on the summary in Section~\ref{sec:training_data}. Our goal in designing this data mixture is to induce video-native spatiotemporal reasoning that (i) supports dense progress estimation under partial observability, and (ii) remains robust when deployed as the sole reward signal for online reinforcement learning. We provide extensive video demonstrations to illustrate what the outputs of our video synthesis approach look like at: \url{https://sole-r1.github.io/}

\subsection{Design Principles}

The data synthesis pipeline is guided by three core principles:

\begin{enumerate}[leftmargin=*]
    \item \textbf{Explicit coverage of partial success and failure states.}
    Online RL policies frequently visit intermediate, incorrect, or regressive states that are underrepresented in expert demonstrations. To prevent optimistic extrapolation and reward hacking, training data must include authentic non-expert behaviors spanning varying degrees of task completion.

    \item \textbf{Temporal locality of supervision.}
    Progress supervision and reasoning are provided at the granularity of individual timesteps, forcing the model to reason about \emph{what changed} between consecutive frames rather than relying on static appearance or final outcomes.

    \item \textbf{Task-structure-aware decomposition.}
    Manipulation tasks admit natural decompositions into subgoals (approach, contact, grasp, transport, release, articulation). We explicitly encode this structure in the non-expert trajectory design so that the model learns reusable progress primitives rather than task-specific heuristics.
\end{enumerate}

\subsection{Non-expert trajectory level taxonomy}

Table~\ref{tab:trajectoryLevels} enumerates the non-expert trajectory levels used in our synthesis pipeline. Each level corresponds to a prefix of a canonical task decomposition, where all preceding subgoals are assumed to be successfully completed, and the failure or termination occurs at the current level.

For example, in pick-and-place tasks, a level-3 trajectory (\emph{Contact}) assumes the robot has successfully approached the object but fails to establish a stable grasp. This structure ensures that trajectories across levels are \emph{semantically ordered} by task progress, enabling consistent supervision of advance versus regress events.

We allocate approximately uniform probability mass across levels \emph{within each task family}. This prevents overrepresentation of near-expert states and ensures that the model is exposed to both early-stage failures (e.g., failure to approach) and late-stage failures (e.g., dropping an object after grasp). Uniform per-level sampling was empirically found to reduce optimistic progress extrapolation compared to expert-heavy mixtures.

\subsection{Simulation-based trajectory synthesis}

For simulation environments (RoboCasa), non-expert trajectories are generated by injecting random action deviations into expert demonstrations, as described in Section~\ref{sec:progress_prediction_data}. Each deviation produces either:
(i) a terminal failure trajectory, or
(ii) a recovery trajectory that interpolates back to a downstream expert state.

Crucially, injected deviations are sampled across the full trajectory, not only near the start. This yields failures that occur after partial task completion (e.g., grasp achieved but object dropped), which are particularly important for teaching the model to distinguish true completion from visually similar near-success states.

Progress supervision is derived from ground-truth simulator geometry and normalized per-trajectory, ensuring that progress values are comparable across different deviation points and trajectory lengths.

\subsection{Real-world video perturbations}

For real-world videos (OXE), where simulator state is unavailable, we synthesize non-expert behavior directly in observation space via temporal perturbations. Temporal reversal windows induce visually plausible regressions (e.g., an object moving away from a goal configuration) while preserving realistic appearance statistics.

By inheriting progress labels from the corresponding expert timestep, reversed segments are explicitly labeled as negative progress, anchoring the model’s reasoning about regression events. This strategy avoids relying on heuristic visual similarity alone and enforces a consistent temporal ordering signal across all real-world data.

\subsection{Grounded chain-of-thought generation}

For each timestep, we generate a chain-of-thought (CoT) explanation describing:
(i) salient visual changes since the previous timestep,
(ii) whether those changes advance or regress the specified goal, and
(iii) the next implied subgoal.

In simulation, CoT traces are initially templated using ground-truth distance and contact signals, then paraphrased using foundation vision-language models to increase linguistic diversity while preserving factual grounding.

In real-world videos, CoT traces are generated directly by a foundation VLM conditioned on the temporal context and the known progress direction. Progress supervision constrains the explanation to remain temporally consistent (e.g., reversed segments must be described as regression), mitigating hallucinated success narratives.

\subsection{Dataset mixture and balancing}

The full dataset mixture is summarized in Table \ref{tab:data_mix}. We group data into three high-level categories:

\begin{itemize}
    \item \textbf{Video-based CoT reasoning and progress prediction}, which directly supervises the SOLE-R1 reward signal.
    \item \textbf{Embodied reasoning and planning}, which provides object-centric and task-structured reasoning grounded in manipulation semantics.
    \item \textbf{General spatial and temporal reasoning}, which supplies broad visual grounding across viewpoints, textures, and domains.
\end{itemize}

Although the raw counts differ substantially across sources, we do not sample proportionally to dataset size. Instead, we apply category-level balancing during SFT (Section~\ref{appendix:training_hyperparam}), ensuring that progress prediction data remains a substantial fraction of each batch without overwhelming general reasoning capabilities.

\subsection{Why authentic non-expert data matters}

Ablation results in Section~\ref{sec:ablations} show that removing authentic non-expert trajectories (while keeping synthetic reversals) significantly increases reward-hacking failures. This suggests that visually realistic but \emph{causally incorrect} states—produced only by real random deviations—are critical for teaching the model distinctions such as:
\vspace{-3mm}
\begin{itemize}
    \item ``near the handle'' vs.\ ``handle actuated'',
    \item ``object aligned'' vs.\ ``object released'',
    \item ``contact'' vs.\ ``stable grasp''.
\end{itemize}

These distinctions are frequently exploited by online RL agents and are difficult to learn from expert-only or temporally monotonic data.

\subsection{Summary}

In summary, the training data synthesis pipeline for SOLE-R1 combines:
(i) structured non-expert trajectory levels aligned with task decompositions,
(ii) simulation-grounded progress supervision,
(iii) observation-space perturbations for real-world video,
and (iv) dense, temporally grounded chain-of-thought reasoning. Together, these design choices produce a dataset that explicitly exposes the model to the states most likely to be encountered—and exploited—during zero-shot online reinforcement learning.

\begin{table}
\caption{Levels for non-expert trajectories and video dataset counts. For each level, we assume that all task components corresponding to the preceding levels have already been successfully completed.}
\label{tab:trajectoryLevels}
\centering
\begin{tabularx}{\textwidth}{l X l}
\toprule
 & \textbf{Non-expert trajectory levels} & \textbf{Percentage} \\
\midrule
Pick-only &
1: Fail to approach \{obj\}& 25\% \\
& 2: Approach \{obj\}& 25\% \\
& 3: Contact \{obj\}& 25\% \\
& 4: Pick up \{obj\} & 25\% \\
\midrule
Pick-and-place &
1: Fail to approach \{obj\}& 14\% \\
& 2: Approach \{obj\}& 14\% \\
& 3: Contact \{obj\}& 14\% \\
& 4: Pick up \{obj\} & 14\% \\
& 5: Keep grasp of \{obj\} & 14\% \\
& 6: Approach placing \{obj\} in \{target\_location\} & 14\% \\
& 7: Place \{obj\} in \{target\_location\} & 14\% \\
\midrule
Open/close doors/drawers & 
1: Fail to approach the door/drawer& 20\% \\
& 2: Approach door/drawer& 20\% \\
& 3: Contact door/drawer& 20\% \\
& 4: Start opening/closing door/drawer& 20\% \\
& 5: Finish opening/closing door/drawer& 20\% \\
\midrule
Turn on/off appliances & 
1: Fail to approach the on/off button/lever& 33\% \\
& 2: Approach the on/off button/lever& 33\% \\
& 3: Successfully adjust the on/off button/lever& 33\% \\
\bottomrule
\end{tabularx}
\end{table}

\newpage
\hfill
\begin{table}[H]
\caption{Dataset mixture}
\label{tab:data_mix}
\centering
\begin{tabular}{l r}
\hline
Data source ID & Total Dataset Count \\
\hline
\multicolumn{2}{l}{\textbf{Video-based CoT Reasoning and Progress Prediction}} \\
\hline
OXE berkeley\_autolab\_ur5 & 4874 \\
OXE berkeley\_fanuc\_manipulation & 2689 \\
OXE bridge & 4602 \\
OXE bridge\_set2 & 32596 \\
OXE bridge\_set3 & 51939 \\
OXE fractal20220817\_data & 9076 \\
OXE fractal20220817\_data\_set2 & 44230 \\
OXE fractal20220817\_data\_set3 & 23362 \\
OXE fractal20220817\_data\_set4 & 35040 \\
OXE jaco\_play & 4106 \\
OXE jaco\_play\_set2 & 2140 \\
OXE nyu\_door\_opening\_surprising\_effectiveness & 4211 \\
OXE ucsd\_kitchen\_dataset\_converted\_externally\_to\_rlds & 1060 \\
RoboCasa CloseDrawer & 79328 \\
RoboCasa OpenDrawer & 80436 \\
RoboCasa OpenSingleDoor & 43601 \\
RoboCasa PnPCabToCounterPickOnly & 22057 \\
RoboCasa PnPCabToCounter & 62951 \\
RoboCasa PnPCounterToCabPickOnly & 27753 \\
RoboCasa PnPCounterToCab & 68311 \\
RoboCasa PnPCounterToMicrowavePickOnly & 25853 \\
RoboCasa PnPCounterToMicrowave & 67014 \\
RoboCasa PnPCounterToSinkPickOnly & 24995 \\
RoboCasa PnPCounterToSink & 55767 \\
RoboCasa PnPCounterToStovePickOnly & 25494 \\
RoboCasa PnPCounterToStove & 61348 \\
RoboCasa PnPMicrowaveToCounterPickOnly & 26006 \\
RoboCasa PnPMicrowaveToCounter & 63625 \\
RoboCasa PnPSinkToCounterPickOnly & 26767 \\
RoboCasa PnPSinkToCounter & 55855 \\
RoboCasa PnPStoveToCounterPickOnly & 26974 \\
RoboCasa PnPStoveToCounter & 68406 \\
RoboCasa TurnOnMicrowave & 52214 \\
RoboCasa TurnOffMicrowave & 51093 \\
\hline
\multicolumn{2}{l}{\textbf{Embodied Reasoning and Planning}} \\
\hline
fsd\_level\_1\_2\_3 & 279739 \\
fsd\_level\_4\_5 & 42074 \\
robovqa & 258280 \\
robo2vlm1 & 678034 \\
robo2vlm1\_reasoning & 4635 \\
ecot\_libero\_all & 504544 \\
ecot\_bridge\_all & 1346416 \\
\hline
\multicolumn{2}{l}{\textbf{General Spatial and Temporal Reasoning}} \\
\hline
spaceom & 806 \\
spacethinker & 12412 \\
spot\_the\_diff & 12562 \\
ssrcot\_llavacot100k & 98522 \\
ssrcot\_spatialqa\_chunk1 & 166639 \\
ssrcot\_spatialqa\_chunk2 & 166639 \\
ssrcot\_spatialqa\_chunk3 & 166640 \\
ssrcot\_visualcot\_chunk1 & 141295 \\
ssrcot\_visualcot\_chunk2 & 141296 \\
ssrcot\_vocot\_chunk1 & 158594 \\
ssrcot\_vocot\_chunk2 & 158594 \\
\hline
\end{tabular}
\end{table}

\section{SOLE-R1 training details}
\label{appendix:training_hyperparam}

This appendix provides full details of the SOLE-R1 model training procedure, expanding on the summary given in  Section \ref{sec:data_synth_model_training}. We describe the backbone architecture, input/output formatting, supervised fine-tuning (SFT) configuration, reinforcement learning with verifiable rewards (RLVR) setup, optimization hyperparameters, and implementation details required for reproducibility.

\subsection{Backbone architecture and initialization}

SOLE-R1 is initialized from \textbf{Qwen3-VL-8B-Instruct}, a vision-language transformer with a ViT-style visual encoder and an autoregressive language decoder. We retain the full multimodal architecture and do not freeze any layers during training. All parameters are updated jointly in both SFT and RLVR stages.

Video inputs are encoded by uniformly sampling frames from the temporal context window described in Section~\ref{sec:video_native_reasoning}. Each frame is resized to $384\times384$ pixels and passed independently through the vision encoder; frame embeddings are concatenated in temporal order and fused with text tokens using the backbone’s native cross-attention mechanism. No explicit temporal positional embeddings beyond the backbone defaults are introduced.

\subsection{Input and output formatting}

All training examples are normalized into a unified multimodal prompt format:
\[
x = [\texttt{<goal>}~g~\texttt{</goal>},\;
      \texttt{<video>}~o_{t-K+1:t}~\texttt{</video>}],
\]
optionally including $\texttt{<prev\_progress>}~p_{t-1}\texttt{</prev\_progress>}$ during training with probability $1-\delta$, where $\delta=0.3$.

The model is trained to emit a structured response:
\[
y = [\texttt{<think>}~m_t~\texttt{</think>},
      \texttt{<answer>}~p_t~\texttt{</answer>}],
\]
where $m_t$ is free-form natural language reasoning and $p_t$ is a scalar progress value represented as a signed integer token in $[-100,100]$. We discretize progress into integer values during training to ensure stable parsing and reward computation.

\subsection{Supervised fine-tuning (SFT)}

\paragraph{Training mixture.}
During SFT, each batch is balanced across three data categories, as stated in Section~\ref{sec:data_synth_model_training}:
(i) foundational spatial and multi-frame temporal reasoning,
(ii) embodied reasoning and planning,
(iii) robot video-based progress prediction.
We enforce approximate per-batch proportions of 40\%, 30\%, and 30\% respectively, using dataset-level sampling weights.

\paragraph{Objective.}
We minimize the standard autoregressive negative log-likelihood loss over the entire output sequence:
\[
\mathcal{L}_{\text{SFT}}
=
-\sum_{t=1}^{|y|}
\log p_\phi(y_t \mid x, y_{<t}).
\]
No auxiliary losses are applied; in particular, we do not separately supervise the progress scalar outside of the language modeling objective.

\paragraph{Optimization.}
SFT is run for a single epoch over the full training mixture.
We use the AdamW optimizer with parameters:
$\beta_1=0.9$, $\beta_2=0.95$, weight decay $=0.1$.
The base learning rate is $1\times10^{-5}$ with cosine decay and a 2,000-step warmup.
Global batch size is 64, distributed across 8 NVIDIA H100 GPUs.

\subsection{Reinforcement learning with verifiable rewards (RLVR)}

After SFT, we further train the model using GRPO-based RLVR, focusing exclusively on the robot video progress prediction dataset.

\paragraph{Sampling.}
For each input query $q$, we sample $G=8$ candidate outputs from the frozen SFT policy $p_{\phi_{\text{old}}}$. Sampling uses temperature $0.7$ and nucleus sampling with $p=0.9$. Candidates that fail to produce a parseable \texttt{<answer>} token are assigned zero reward.


\paragraph{GRPO parameters.}
We use $\epsilon=0.2$ for ratio clipping and $\beta=0.01$ for KL regularization against the SFT reference model.
RLVR is run for 5,000 steps over the progress dataset, with a learning rate of $1\times10^{-6}$.

\subsection{Compute and infrastructure}

Training is conducted on clusters of NVIDIA H100 GPUs.
SFT requires approximately 1 week on a node of 8 GPUs, while RLVR requires an additional 2 days.

\subsection{Reproducibility details}

We release:
(i) the full training mixture with dataset identifiers and sampling weights,
(ii) exact prompt templates and output parsers,
(iii) SFT and RLVR training scripts with configuration files,
(iv) final model checkpoints after SFT and RLVR.
This enables exact reproduction of the training procedure.





\section{General spatial and vision-language reasoning}
In Table \ref{tab:ssr_improvement}, we show that SOLE-R1 achieves improved performance on three spatial and general visual reasoning benchmarks relative to the backbone models and the strong SSR spatial reasoning model. The benchmarks include: SpatialBench \cite{xu2025spatialbench}, SSRBench \cite{liu2025ssr}, and CV-Bench \cite{zhu2025cvbench}. We observe consistent gains across all three benchmarks. On SpatialBench, SOLE-R1 improves over the SSR 7B model by +2.4 points and over the backbone models by a larger margin. On SSRBench, SOLE-R1 achieves the strongest performance in both the General and Spatial subsets, outperforming SSR 7B by +3.5 and +3.1 points, respectively, indicating improved generalization and spatial reasoning capability. 

\begin{table*}[ht]
\centering
\caption{Performance improvement of SOLE-R1 compared to SSR 7B model and the backbone model on SpatialBench, SSRBench, and CV-Bench.}
\label{tab:ssr_improvement}
\setlength{\tabcolsep}{6pt}
\begin{tabular}{lccccccccc}
\toprule
\multirow{2}{*}{} &
\multicolumn{1}{c}{ } &
\multirow{2}{*}{SpatialBench} &
\multicolumn{2}{c}{SSRBench} &
\multirow{2}{*}{CV-Bench} \\
\cmidrule(lr){4-5}\cmidrule(lr){7-8}
Method & Size  & & General & Spatial & \\
\midrule
Qwen2.5-VL   & 7B & 64.7 & 69.4 & 53.6 & 73.0  \\
Qwen3-VL   & 8B & 65.3 & 73.1 & 57.2 & 73.7 \\
\text{SSR}  & 7B & 67.0 & 82.1  & 76.1  & 73.3   \\
\rowcolor{black!8}
\textbf{SOLE-R1 (Ours)}  & 8B & 69.4  & 85.6 & 79.2  & 74.4   \\
\bottomrule
\end{tabular}
\end{table*}

\begin{table}[H]
\caption{Value-Order-Correlation analysis (as done in on \cite{ma2024vision})
comparing SOLE-R1 to GVL on 1,000 videos from OpenX Embodiment datasets.}
\label{tab:vocScores}
\centering
%
\begin{tabularx}{\textwidth}{X l l}
\toprule
Dataset & GVL & SOLE-R1 \\
\midrule
\texttt{utokyo\_xarm\_pick\_and\_place\_converted\_externally\_to\_rlds} & 0.74 & 0.92 \\
\texttt{utokyo\_xarm\_bimanual\_converted\_externally\_to\_rlds} & 0.83 & 0.86 \\
\texttt{nyu\_door\_opening\_surprising\_effectiveness} & 0.69 & 0.88 \\
\texttt{berkeley\_autolab\_ur5} & 0.77 & 0.85 \\
\texttt{utokyo\_pr2\_tabletop\_manipulation\_converted\_externally\_to\_rlds} & 0.65 & 0.87 \\
\texttt{maniskill\_dataset\_converted\_externally\_to\_rlds} & 0.72 & 0.84 \\
\texttt{utokyo\_pr2\_opening\_fridge\_converted\_externally\_to\_rlds} & 0.79 & 0.88 \\
\cellcolor{yellow}\texttt{\texttt{fractal20220817\_data}} & \cellcolor{yellow}0.77 & \cellcolor{yellow}0.84 \\
\texttt{iamlab\_cmu\_pickup\_insert\_converted\_externally\_to\_rlds} & 0.56 & 0.81 \\
\texttt{toto} & 0.55 & 0.78 \\
\texttt{ucsd\_kitchen\_dataset\_converted\_externally\_to\_rlds} & 0.53 & 0.79 \\
\texttt{utaustin\_mutex} & 0.60 & 0.75 \\
\texttt{asu\_table\_top\_converted\_externally\_to\_rlds} & 0.46 & 0.74 \\
\texttt{austin\_sirius\_dataset\_converted\_externally\_to\_rlds} & 0.57 & 0.76 \\
\cellcolor{yellow}\texttt{\texttt{dobbe}} & \cellcolor{yellow}0.51 & \cellcolor{yellow}0.73 \\
\texttt{berkeley\_cable\_routing} & 0.49 & 0.75 \\
\texttt{berkeley\_rpt\_converted\_externally\_to\_rlds} & 0.48 & 0.73 \\
\texttt{viola} & 0.44 & 0.77 \\
\texttt{fmb} & 0.58 & 0.74 \\
\texttt{austin\_buds\_dataset\_converted\_externally\_to\_rlds} & 0.39 & 0.71 \\
\texttt{usc\_cloth\_sim\_converted\_externally\_to\_rlds} & 0.46 & 0.70 \\
\cellcolor{yellow}\texttt{\texttt{bridge}} & \cellcolor{yellow}0.48 & \cellcolor{yellow}0.72 \\
\texttt{jaco\_play} & 0.42 & 0.69 \\

\texttt{stanford\_hydra\_dataset\_converted\_externally\_to\_rlds} & 0.37 & 0.71 \\
\texttt{bc\_z} & 0.41 & 0.73 \\
\texttt{berkeley\_mvp\_converted\_externally\_to\_rlds} & 0.34 & 0.67 \\
\texttt{cmu\_stretch} & 0.31 & 0.66 \\
\texttt{tokyo\_u\_lsmo\_converted\_externally\_to\_rlds} & 0.35 & 0.65 \\
\texttt{berkeley\_fanuc\_manipulation} & 0.27 & 0.63 \\
\texttt{roboturk} & 0.32 & 0.62 \\
\texttt{ucsd\_pick\_and\_place\_dataset\_converted\_externally\_to\_rlds} & 0.26 & 0.59 \\
\texttt{dlr\_edan\_shared\_control\_converted\_externally\_to\_rlds} & 0.15 & 0.57 \\
\texttt{dlr\_sara\_pour\_converted\_externally\_to\_rlds} & 0.10 & 0.55 \\
\texttt{droid} & 0.04 & 0.53 \\
\texttt{taco\_play} & 0.09 & 0.56 \\
\texttt{stanford\_robocook\_converted\_externally\_to\_rlds} & -0.02 & 0.51 \\
\texttt{imperialcollege\_sawyer\_wrist\_cam} & -0.01 & 0.46 \\
\texttt{kaist\_nonprehensile\_converted\_externally\_to\_rlds} & -0.05 & 0.41 \\
\texttt{austin\_sailor\_dataset\_converted\_externally\_to\_rlds} & -0.07 & 0.38 \\
\cellcolor{white}\texttt{\texttt{kuka}} & \cellcolor{white}0.28 & \cellcolor{white}0.35 \\
\texttt{cmu\_play\_fusion} & -0.08 & 0.33 \\
\texttt{stanford\_kuka\_multimodal\_dataset\_converted\_externally\_to\_rlds} & -0.14 & 0.30 \\
\texttt{nyu\_franka\_play\_dataset\_converted\_externally\_to\_rlds} & -0.16 & 0.29 \\
\texttt{stanford\_mask\_vit\_converted\_externally\_to\_rlds} & -0.11 & 0.27 \\
\texttt{uiuc\_d3field} & -0.20 & 0.18 \\
\cellcolor{white}\texttt{\texttt{robo\_net}} & \cellcolor{white}-0.24 & \cellcolor{white}0.12 \\
\texttt{columbia\_cairlab\_pusht\_real} & -0.26 & 0.09 \\
\texttt{dlr\_sara\_grid\_clamp\_converted\_externally\_to\_rlds} & -0.28 & -0.01 \\
\texttt{cmu\_franka\_exploration\_dataset\_converted\_externally\_to\_rlds} & -0.25 & -0.04 \\

\bottomrule
\end{tabularx}
\end{table}

\section{VLA RL Steering}
\label{appendix:vla_steering}

\begin{figure}[H]
    \centering
    \includegraphics[width=\linewidth]{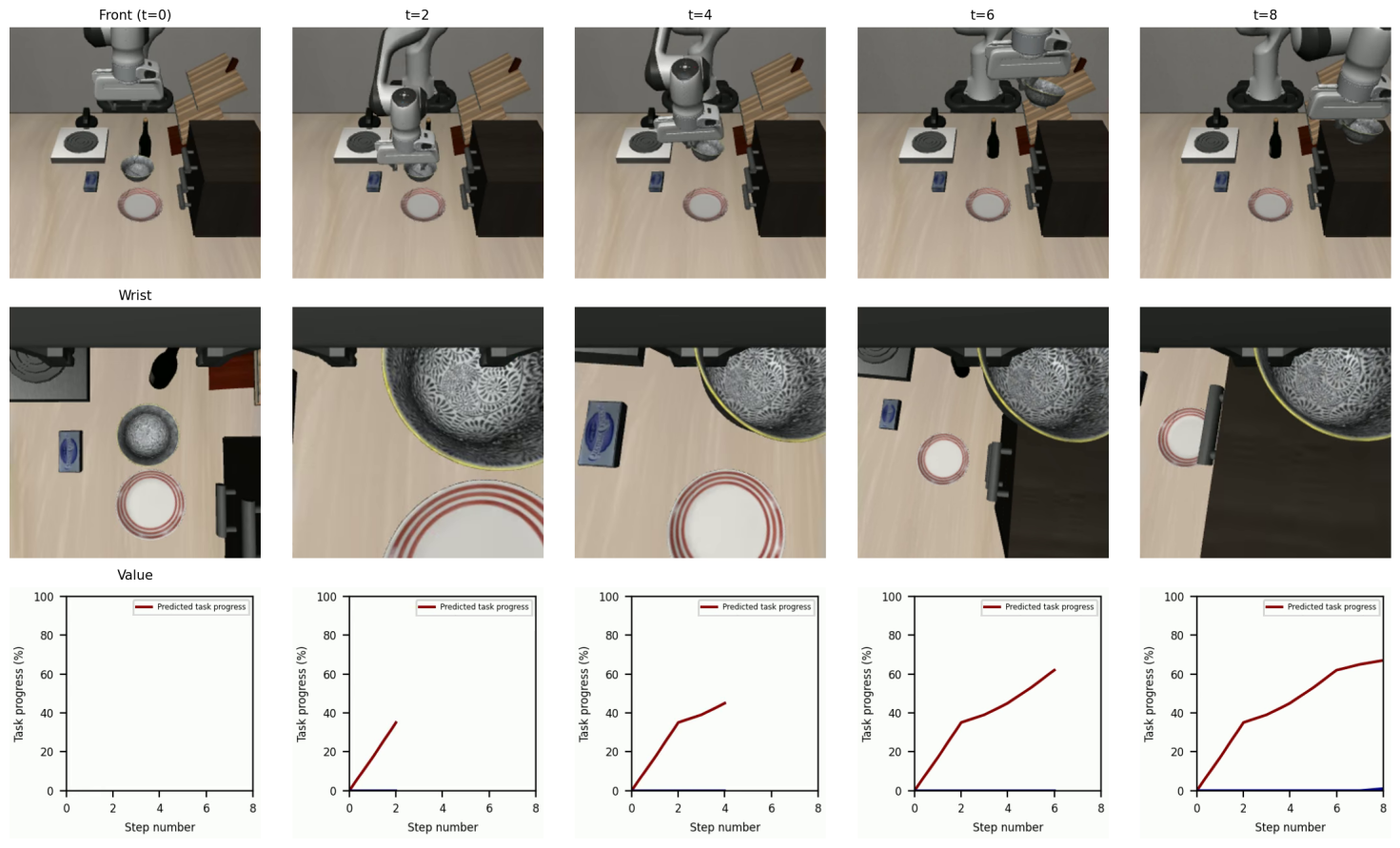}
    \includegraphics[width=\linewidth]{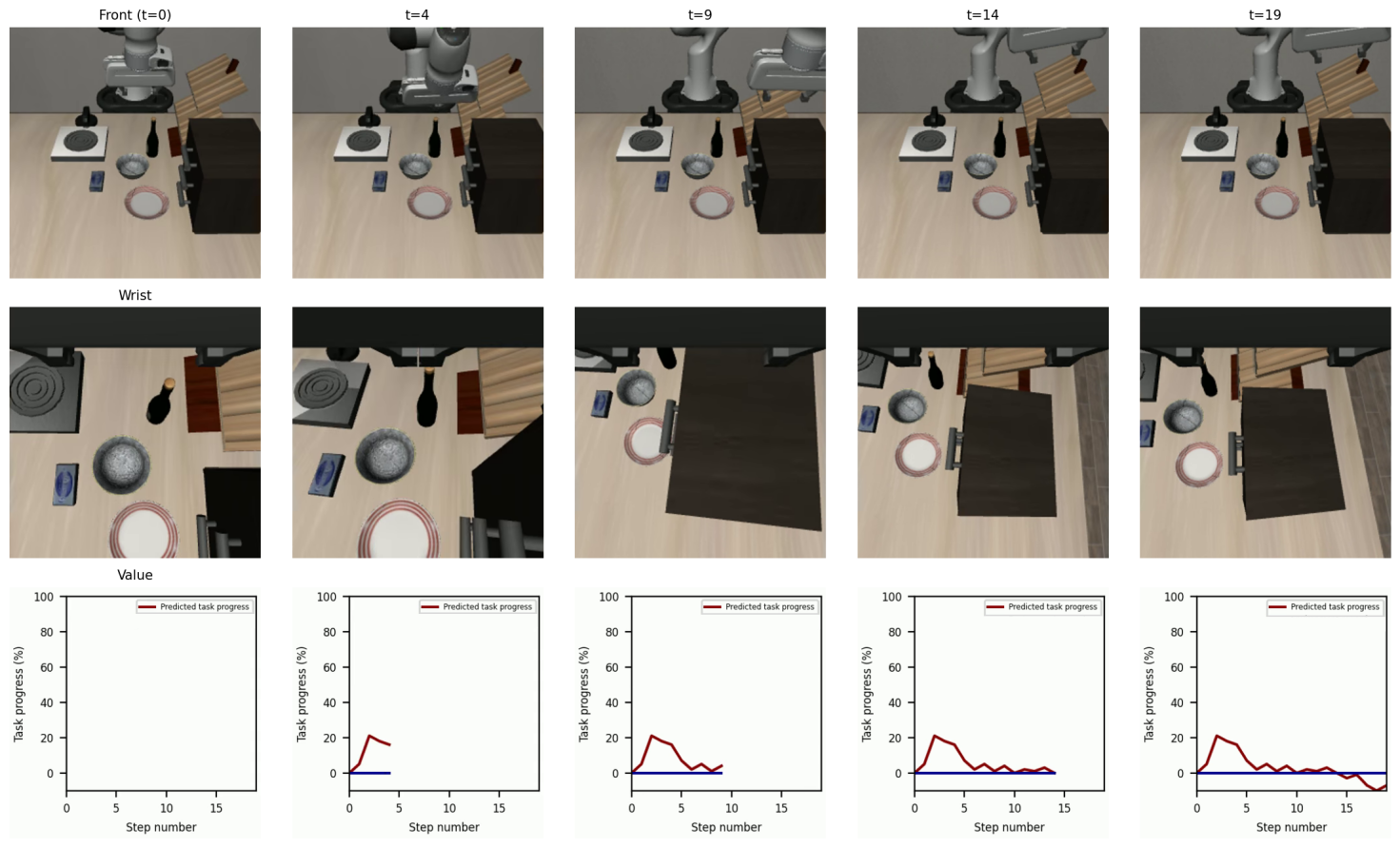}
    \caption{Example of a successful (top) and failed (bottom) manipulation trajectory generated by SmolVLA in Libero. We predict and plot (red) SOLE-R1 task progress predictions based on the prompt ``put the bowl on the cabinet''.}
    \label{fig:smolvla_1}
\end{figure}

\begin{figure}[H]
    \centering
    \includegraphics[width=\linewidth]{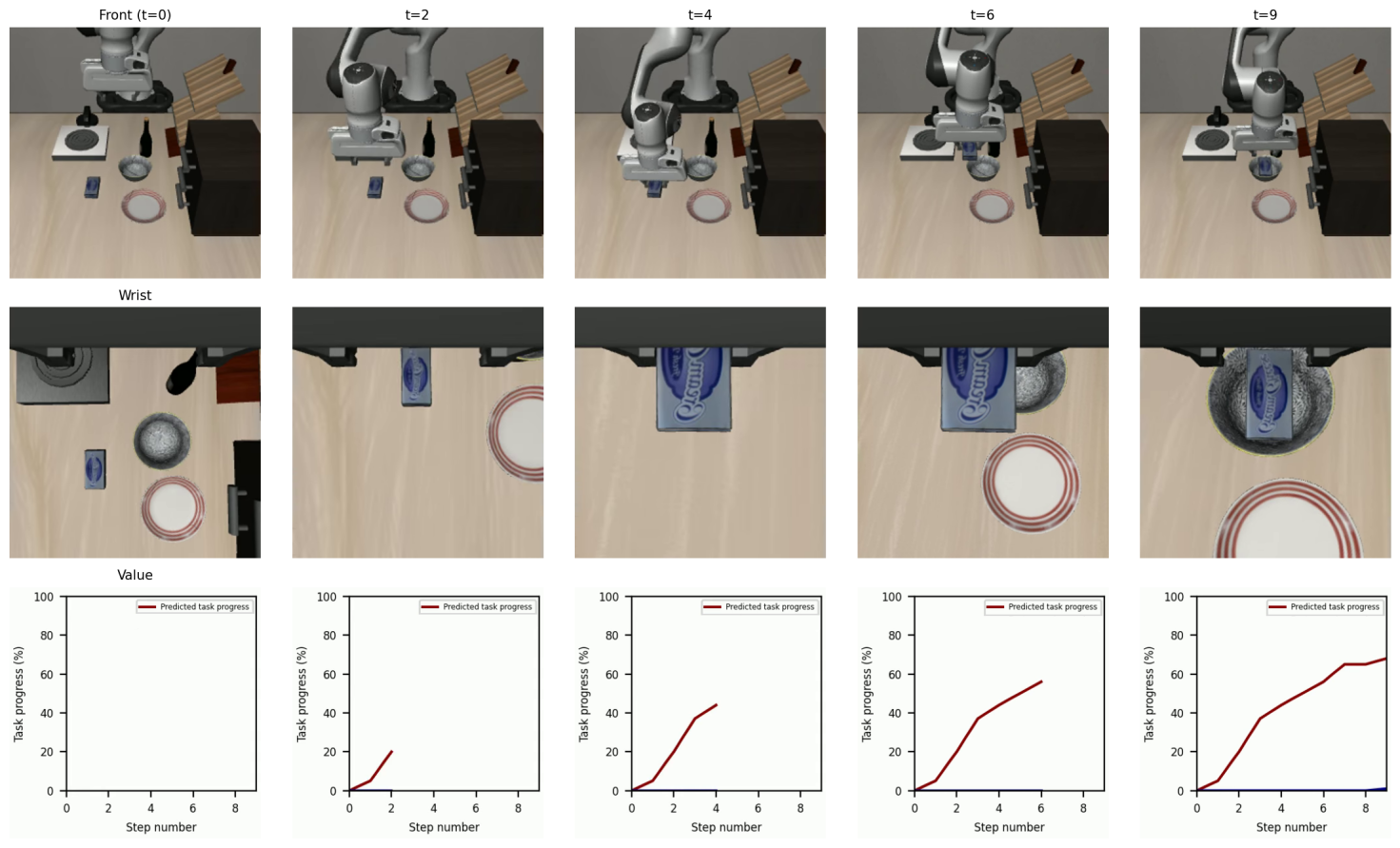}
    \includegraphics[width=\linewidth]{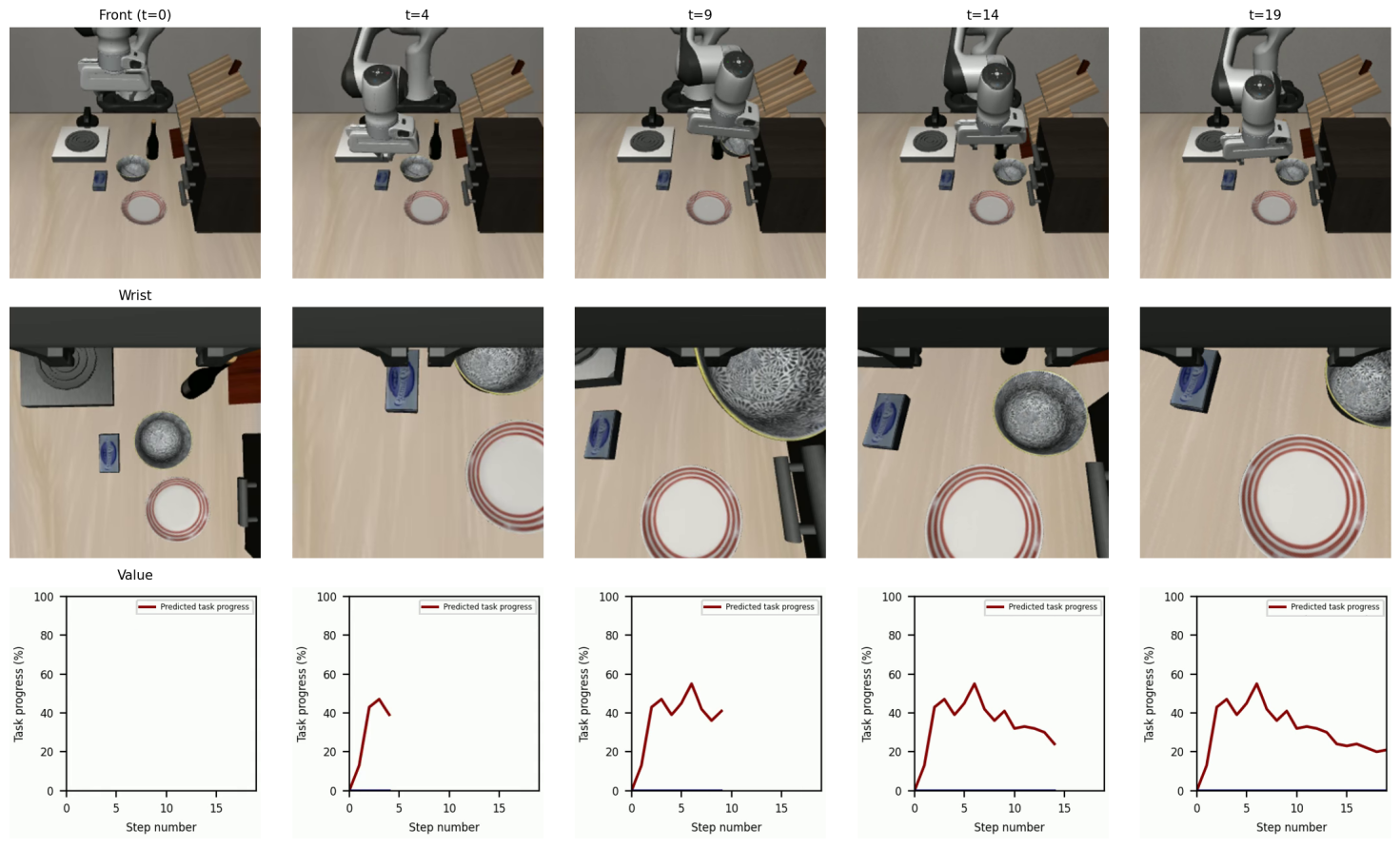}
    \caption{Example of a successful (top) and failed (bottom) manipulation trajectory generated by SmolVLA in Libero. We predict and plot (red) SOLE-R1 task progress predictions based on the prompt ``put the cream cheese in the bowl''.}
    \label{fig:smolvla_2}
\end{figure}

\section{Details of zero-shot scaling experiments}
\label{appendix:zero_shot_scaling}

This appendix provides additional details for the \emph{Zero-shot Scaling} experiment reported in Section~\ref{sec:scaling}. The goal of this experiment is to isolate how \emph{training-task diversity} in the video-based progress supervision affects downstream zero-shot online RL performance, while holding model architecture, optimization, and total training compute fixed.

\subsection{Training task-type taxonomy}

To study scaling with respect to task diversity, we group all robot-video progress supervision into a set of \textbf{seven canonical task types}. Each task type corresponds to a distinct manipulation primitive with characteristic spatiotemporal progress structure, contact dynamics, and failure modes. The full list is:

\vspace{-2mm}
\begin{enumerate}[leftmargin=*]
    \item \textbf{PickOnly}: Grasping and lifting a single object from a support surface, without placement.
    \item \textbf{Pick-and-Place (Pick $\rightarrow$ Counter)}: Picking an object from a receptacle (e.g., cabinet, microwave, sink, stove) and placing it onto a counter or open surface.
    \item \textbf{Pick-and-Place (Counter $\rightarrow$ Receptacle)}: Picking an object from a counter and placing it into a target receptacle (e.g., cabinet, microwave, sink).
    \item \textbf{Open/Close Drawer}: Articulated manipulation involving sliding drawers, including approach, contact, actuation, and completion states.
    \item \textbf{Open/Close Door}: Articulated manipulation of hinged doors, requiring handle interaction and rotational motion.
    \item \textbf{Button-Press}: Discrete actuation of buttons or switches (e.g., microwave, coffee machine), characterized by brief contact events and state changes with minimal visual motion.
    \item \textbf{Lever / Knob Actuation}: Continuous rotational or pulling actions on levers or knobs, including turning appliances on or off.
\end{enumerate}


\subsection{Scaling protocol}

We construct a sequence of SOLE-R1 variants by progressively increasing the number of task types included in the \emph{video-based progress prediction} portion of the training data. Concretely:

\vspace{-2mm}
\begin{itemize}
    \item All models share the same backbone (Qwen3-VL-8B-Instruct), training recipe (SFT + RLVR), optimizer settings, and total number of training steps.
    \item Foundational spatial reasoning and multi-frame temporal reasoning data are held constant across all variants.
    \item Only the \emph{set of task types} used to generate robot-video progress supervision is varied.
\end{itemize}


\subsection{Evaluation metric}

Each trained model is evaluated on the full zero-shot online RL benchmark described in Section~\ref{sec:zero_shot_rl}, comprising tasks \emph{not used} for training progress supervision. For each model, we report the number of downstream tasks achieving success rates above fixed thresholds (e.g., 10\%, 30\%, 50\%).
Figure~\ref{res_fig4} plots the number of tasks solved above threshold as a function of training task-type diversity.

\end{document}